\documentclass{article}

\PassOptionsToPackage{numbers,sort}{natbib}
\PassOptionsToPackage{table,xcdraw}{xcolor}

\usepackage{iclr2027_conference,times}

\iclrfinalcopy

\usepackage[utf8]{inputenc}
\usepackage[T1]{fontenc}

\usepackage{amsmath,amssymb,amsfonts}
\usepackage{amsthm}
\usepackage{nicefrac}
\usepackage{dsfont}
\usepackage{microtype}

\usepackage{graphicx}
\usepackage{booktabs}
\usepackage{multirow}
\usepackage{tabularx}
\usepackage{adjustbox}
\usepackage{makecell}
\usepackage{rotating}
\usepackage{float}
\usepackage{wrapfig}
\usepackage{subcaption}

\usepackage{algorithm}
\usepackage{algpseudocode}
\usepackage{algorithmicx}

\usepackage{enumitem}
\usepackage{xspace}
\usepackage{listings}
\usepackage{xcolor}
\usepackage[most]{tcolorbox}
\usepackage{lineno}

\usepackage{url}
\usepackage{hyperref}

\usepackage{amsmath,amsfonts,bm}

\def\figref#1{figure~\ref{#1}}

\def\secref#1{section~\ref{#1}}

\def\eqref#1{equation~\ref{#1}}

\def\algref#1{algorithm~\ref{#1}}

\def\1{\bm{1}}

\DeclareMathAlphabet{\mathsfit}{\encodingdefault}{\sfdefault}{m}{sl}
\SetMathAlphabet{\mathsfit}{bold}{\encodingdefault}{\sfdefault}{bx}{n}

\setcitestyle{numbers,square,comma}

\hypersetup{
    colorlinks=true,
    linkcolor=red,
    citecolor=cyan,
    filecolor=magenta,
    urlcolor=magenta
}

\usepackage{fontawesome5}

\newcommand{\githubrepo}[1]{%
\href{#1}{\faGithub\ \texttt{Code}}%
}

\newcommand{\tabref}[1]{Table~\ref{#1}}
\newcommand{\eqnref}[1]{Eq.~(\ref{#1})}
\newcommand{\appref}[1]{Appendix~\ref{#1}}

\providecommand{\algref}[1]{Algorithm~\ref{#1}}

\newcommand{\ours}{\textsc{SPD}\xspace}

\definecolor{upcolor}{HTML}{F28B71}
\definecolor{downcolor}{HTML}{48C1B6}
\definecolor{bestbg}{HTML}{DEE6F0}
\definecolor{tagbg}{HTML}{B4B4B4}

\definecolor{softpurple}{RGB}{236,229,246}
\definecolor{softorange}{RGB}{248,231,213}
\definecolor{softrose}{RGB}{222,239,235}
\definecolor{softred}{RGB}{248,220,220}

\newcommand{\hlg}[1]{\cellcolor{softpurple}#1}
\newcommand{\hlb}[1]{\cellcolor{softorange}#1}
\newcommand{\hlr}[1]{\cellcolor{softrose}#1}

\newcommand{\res}[2]{%
  #1_{\scriptscriptstyle \pm #2}%
}

\newcommand{\purplebox}[1]{%
  {\setlength{\fboxsep}{1pt}%
   \colorbox{softpurple}{#1}}%
}

\newcommand{\orangebox}[1]{%
  {\setlength{\fboxsep}{1pt}%
   \colorbox{softorange}{#1}}%
}

\newcommand{\rosebox}[1]{%
  {\setlength{\fboxsep}{1pt}%
   \colorbox{softrose}{#1}}%
}

\title{
Self-Policy Distillation \\
via Capability-Selective Subspace Projection
}

\author{
Guangya Hao\textsuperscript{1}
\quad
Hanxue Liang\textsuperscript{1}
\quad
Yitong Shang\textsuperscript{1,2}
\quad
Yunbo Long\textsuperscript{1}
\quad
Zhuokai Zhao\textsuperscript{3}
\\[0.4em]
\textsuperscript{1}University of Cambridge
\quad
\textsuperscript{2}HKUST
\quad
\textsuperscript{3}University of Chicago
\\[0.6em]
{\normalfont\small
Correspondence:
\texttt{gh540@cam.ac.uk} and
\texttt{zhuokai@uchicago.edu}
}
}

\begin{document}


\maketitle

\fancyhead[L]{Preprint}


\begin{abstract}

Self-distillation bootstraps large language models (LLMs) by training on their own generations. 
However, existing methods either rely on external signals to curate self-generated outputs (e.g., correctness filtering, execution feedback, and reward search), which are costly and unavailable for the best-performing frontier models, or skip curation entirely and train on all raw outputs, an approach that is often domain-specific and hard to generalize.
Both also share a deeper weakness that self-generated outputs entangle task-relevant capability with others, such as stylistic patterns, formatting artifacts, and model-specific errors, diluting the signal for the specific capability one aims to improve. 
In this paper, we propose \textbf{ Self-Policy Distillation (\ours)}, which achieves generalizable, capability selective with only minimal external signal (\(<5\%\) labeled data in each domain).
Specifically, \ours extracts a low-rank capability subspace from the model's own gradients on correctness-defining tokens, projects key-value (KV) activations into this subspace during self-generation, and fine-tunes on the resulting raw outputs with standard next-token prediction loss.
Through extensive experiments across code generation, mathematical reasoning, and multiple-choice QA, we show that \ours achieves up to \(13\%\) improvement over state-of-the-art self-distillation methods and up to \(16.3\%\) improvement over pre-trained baselines.
Notably, \ours demonstrates superior generalizability, achieving \(15.7\%\) better performance under out-of-domain generalization settings.

\end{abstract}

\begin{center}
    \vspace{-0.5em}
    \githubrepo{https://github.com/gh540-svg/SPD}
    \vspace{-0.5em}
\end{center}


\begin{figure}[htp]
    \centering
    \includegraphics[width=\linewidth]{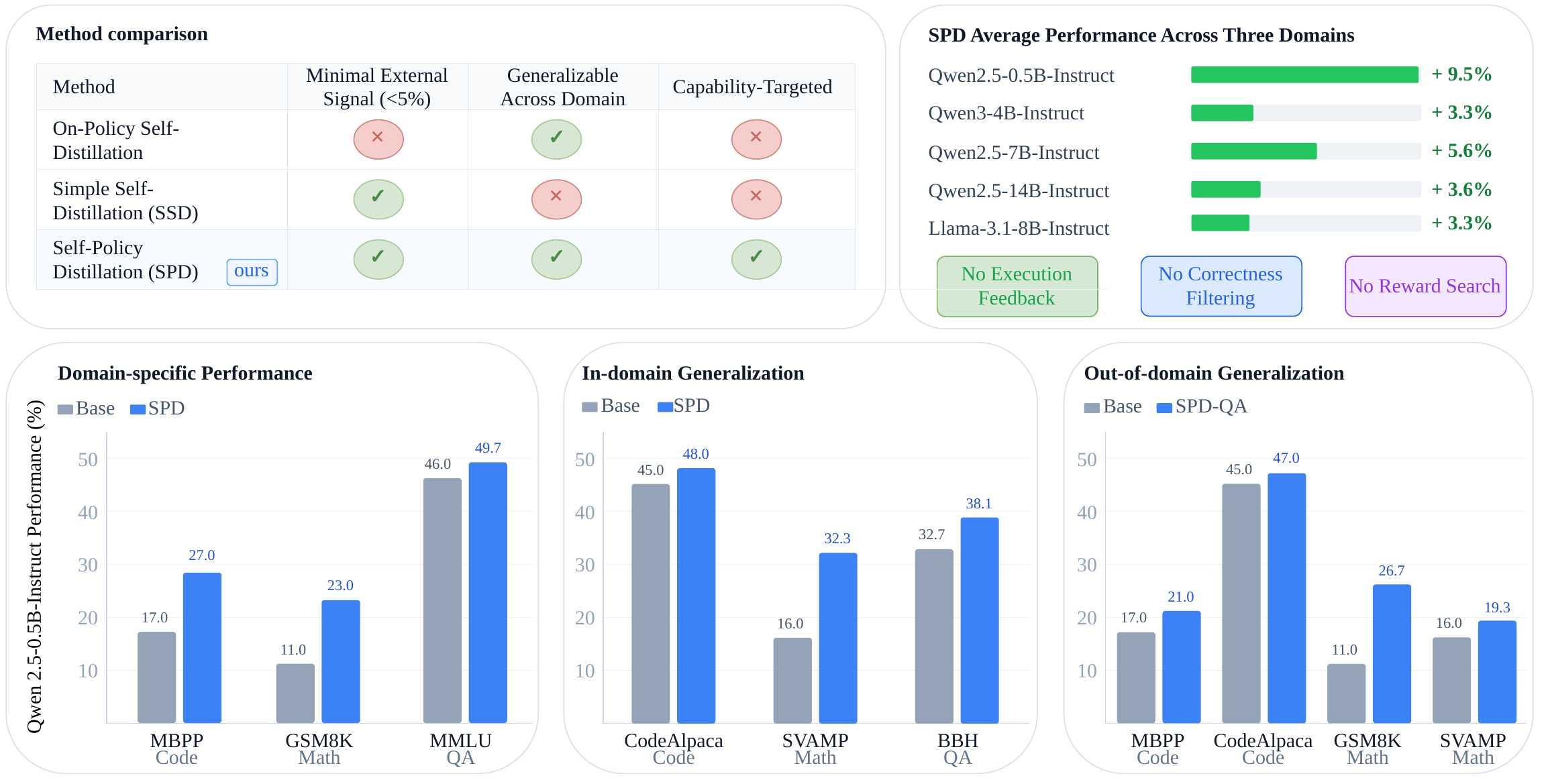}  
    \vspace{-0.2in}
    \caption{\ours comparisons and performance summary. Top left: comparison with existing self-distillation methods across three key axes. Top right: average improvement of \ours over the base model for each of the five LLM backbones, computed across three capability domains and six datasets. Bottom: \ours improves base model performance across domain-specific benchmarks, in-domain transfer, and out-domain transfer settings. All metrics in the figure are reported as accuracy (\%).}
    \vspace{-0.2in}
    \label{fig:intro_teaser}
    \vspace{-0.1in}
\end{figure}
\section{Introduction}
Self-distillation has emerged as a powerful paradigm for improving large language models (LLMs) by training them on their own generations~\citep{zelikman2024star,singh2023beyond,zhang2026ssd}. 
Existing methods typically do not use raw self-generated outputs directly,  because doing so can reinforce the model's existing errors and stylistic artifact, leading to results in poor performance~\citep{li2026rethinking,dai2025capture,lippmann2025style,chen2025unveiling}. 
Instead, one line of works relies on external signals to curate or guide self-generated data, such as correctness filtering~\citep{zelikman2024star}, verifier-based selection~\citep{hosseini2024v}, consistency-driven rationale evaluation~\citep{lee2025self}, or reward-guided search~\citep{zhang2024rest}.
While effective, these signals introduce additional cost and task-specific infrastructure. 
In addition, they may be unavailable for the strongest frontier models, where more reliable labels, verifiers, or interactive feedback are often hard to obtain~\citep{Penaloza2026PrivilegedID,song2026survey}.
Another line of work skips explicit curation entirely and trains directly on raw self-generated outputs after internal filtering.
Simple self-distillation (SSD)~\citep{zhang2026ssd} shows that this simple strategy can improve model performance without using a verifier, reward model, or reinforcement learning (RL).
However, SSD is designed for code generation, and its generalization beyond the code domain remains unclear.

Additionally, both lines of work share a deeper weakness: self-generated outputs are a mixed source of supervision.
They contain not only the task-relevant capability signal we aim to preserve, but also model-specific errors and noisy artifacts that can hinder capability transfer~\citep{li2026rethinking,dai2025capture}, as well as stylistic patterns, rationale formats, and other non-essential reasoning details that may dilute the signal for the specific capability being improved~\citep{lippmann2025style,chen2025unveiling}. 
\figref{fig:intro_teaser} (top-left) summarizes that  existing self-distillation paradigms each fall short on two of these key axes.
These bottlenecks raise a central question:
%
%
\begin{center}
\textit{Can self-distillation be made with only minimal external signals by \textbf{internally steering} self-generated outputs to \textbf{generalize} across domains?}
\end{center}

To address this question, we propose \textbf{Self-Policy Distillation (\ours)}, a generalizable and capability-selective framework that requires only minimal external signals. 
\ours consists of two phases. 
In the first phase, \ours uses a small calibration set, accounting for less than 5\% of the training data in each domain, to extract a low-rank capability subspace from the model's own gradients.
These gradients are computed using a loss defined only on \textit{correctness-defining} token positions.
In practice, for each calibration example, we compute a loss only over the task-relevant answer span.
We then collect the gradients with respect to the key and value activations at selected layers.
Finally, we perform singular value decomposition (SVD) on these gradients to obtain a low-rank projection subspace for the key ($K$) and value ($V$) activations, which are used to construct the \textit{projection hooks} in second phase. 

In the second phase, we use this capability subspace to steer self-generation.
Specifically, we register projection hooks on the corresponding KV activations during decoding.
These hooks do not modify the model parameters.
Instead, they filter intermediate representations so that the model's self-generated outputs are biased toward the target capability.
Using the hooked model, we generate one raw completion for each training prompt and construct a self-generated corpus for distillation.
After data generation, all hooks are removed, and the original model is fine-tuned on the resulting prompt--completion pairs with the standard loss.
Therefore, \ours remains a fully self-contained self-distillation pipeline that requires only minimal external signals, while the projection hooks help selectively preserve target capability signals and filter out task-irrelevant noise.

We evaluate \ours across three capability domains --- code generation, mathematical reasoning, and multiple-choice question-answering (QA) --- under two transfer settings that test whether learned gains generalize within a capability domain (\textbf{in-domain}) and beyond it (\textbf{out-of-domain}).
As summarized in \figref{fig:intro_teaser}, \ours consistently improves over the base model across five backbones (top-right), three source benchmarks (bottom-left), and both transfer settings (bottom-middle and bottom-right).
%
%
Specifically, \ours achieves up to $13\%$ improvement over state-of-the-art self-distillation approaches and up to $16.3\%$ improvement over pre-trained baselines under the in-domain setting. 
Notably, \ours also demonstrates strong generalization, achieving up to $15.7\%$ better performance under the out-of-domain setting. 

To summarize, our contributions are outlined below:
\begin{itemize}[topsep=0pt,leftmargin=*,noitemsep]

    \item  We propose \textbf{Self-Policy Distillation (\ours)}, a new self-distillation paradigm that trains on selective self-generated outputs.
    \ours requires only minimal external signals and generalizes across domains by selectively preserving target capability signals during self-generation.

    \item We define \textit{correctness spans}, which are token positions that directly determine task success, and apply a correctness-aligned loss only on these positions. 
    SVD on the resulting K/V activation gradients yields a low-rank capability subspace that steers self-generation toward the target capability.
    %
    

    \item Across code generation, mathematical reasoning, and multiple-choice QA, \ours consistently outperforms state-of-the-art self-distillation methods under both in-domain and out-of-domain generalization settings.
    
   
\end{itemize}


\section{Related Work}
\label{sec:related}

\paragraph{Distillation.}
Early off-policy distillation methods mainly focus on transferring information from a stronger teacher to a student model, either through softened output distributions~\citep{hinton2015distilling} or sequence-level supervision~\citep{kim2016sequence}.
However, such off-policy training introduces a train-inference distribution mismatch that leads to compounding errors at test time~\citep{chen2025retaining}.
Later, on-policy distillation (OPD)~\citep{agarwal2024policy, boizard2024towards} mitigates this mismatch by using student rollouts while receiving per-token supervision from the teacher.
However, it requires repeated student rollouts and heavy teacher supervision, making it costly and hard to scale.
Self-distillation mitigates these limitations by replacing the external teacher with supervision derived from the model itself~\citep{askell2021general, zhang2024rest, zhang2026ssd}.
However, directly using self-generated outputs for training can reinforce the model's existing errors and stylistic artifact, leading to poor performance~\citep{li2026rethinking,dai2025capture,lippmann2025style,chen2025unveiling}. 
Therefore, improving the quality of self-generated data is critical.
Existing methods rely on external signals to curate self-generated data, such as correctness filtering~\citep{zelikman2024star}, consistency or execution-based feedback~\citep{lee2025self, gehring2024rlef}, or reward-guided search~\citep{zhang2024rest}. 
While effective, these signals introduce additional cost,task-specific infrastructure and unavailable for the best-performing frontier models.
More recently, SSD~\citep{zhang2026ssd} shows that a LLM can improve by training directly on its own raw outputs, without a verifier, a teacher model, or RL.
However, SSD is designed for code generation and trained on domain-specific synthetic code data, leaving its generalization beyond the code domain unclear. 
In contrast, our work keeps the self-distillation setting with only minimal external signals, while making self-generated data capability-selective and  generalizable across domains. 
A more detailed discussion comparing \ours (ours) with existing methods is in \appref{sec:distillation_paradigms}.

\paragraph{Representation-level control.}
Recent work suggests that meaningful reasoning behavior and task-relevant information can often be compressed, manipulated, or steered in latent or KV-space representations rather than only at the output level~\citep{hao2024training, shen2025codi, chari2025kv, wang2025system}.
In parallel, prior work on distillation has shown that intermediate representations can provide richer and more targeted supervision than output logits alone~\citep{romero2015fitnets, sun2019patient}.
Our method combines these two perspectives in a teacher-free self-distillation setting.
Specifically, rather than aligning a student to an external teacher's hidden states, we extract a capability-selective subspace from the student's own gradients and use it to steer self-generation, producing selective raw data for more useful self-supervision.
%


\section{Method}\label{sec:method}
\begin{figure}[t]  
    \centering
    \includegraphics[width=1\linewidth]{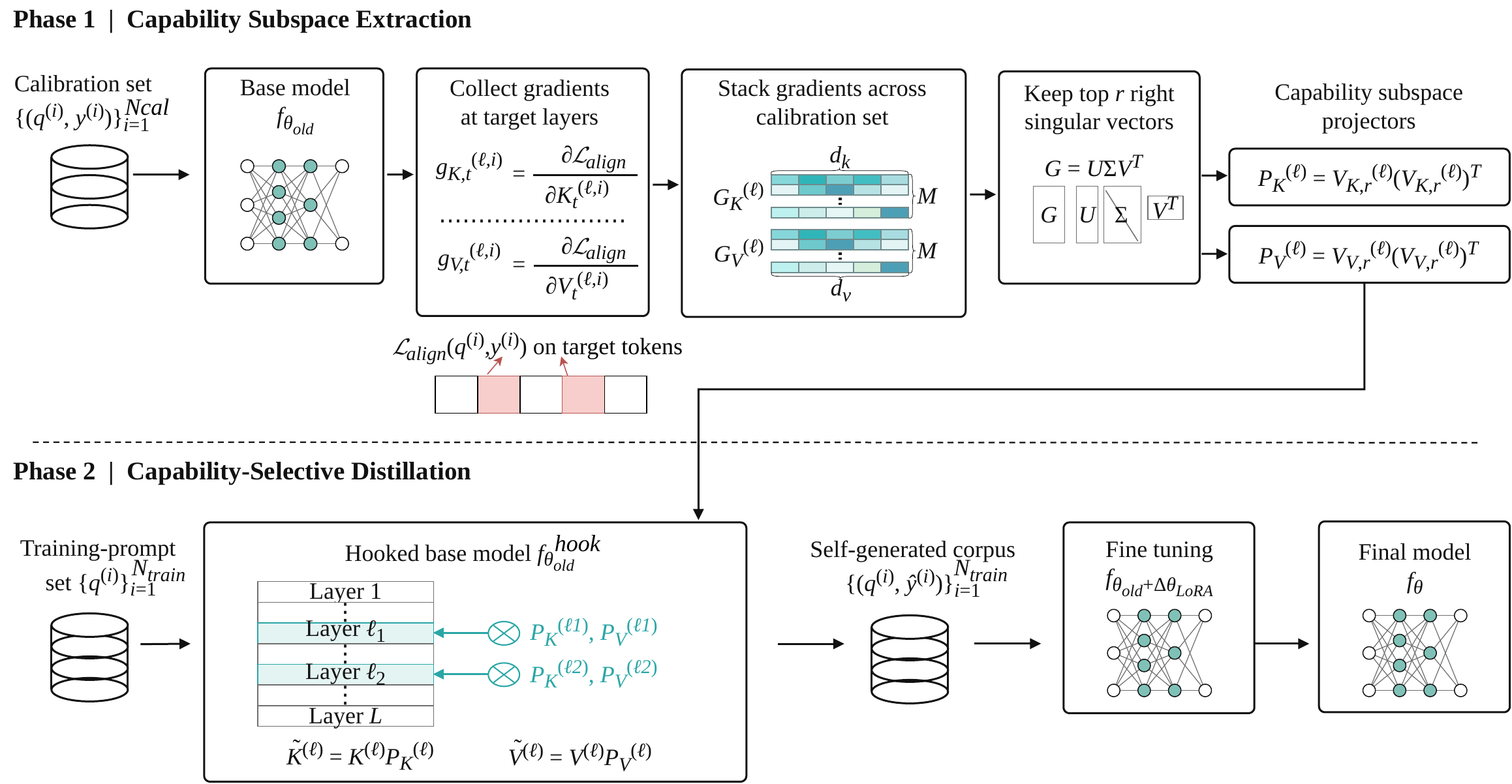}  
    \caption{Overview of \ours. \ours operates in two phases: Phase~1 (top) (\secref{sec:method:subspace}) extracts low-rank K/V capability subspaces from gradients computed on a small calibration set using correctness-aligned loss, and Phase~2 (bottom) (\secref{sec:method:hook}) uses these subspaces as projection hooks to steer self-generation without modifying model parameters. The hooked model produces raw completions, after which the hooks are removed and the original model is fine-tuned on the resulting prompt--completion pairs using standard next-token prediction loss.}
    \label{fig:calibration}
\end{figure}
In this section, we illustrate \textbf{self-policy distillation (\ours)}, a capability-selective framework that requires only minimal external signals. 
On a high level, \ours operates in two phases: Phase~1 (\secref{sec:method:subspace}) extracts low-rank K/V capability subspaces from gradients computed on a small calibration set using correctness-aligned loss, and Phase~2 (\secref{sec:method:hook}) uses these subspaces as projection hooks to steer self-generation without modifying model parameters.
The hooked model produces raw completions, after which the hooks are removed and the original model is fine-tuned on the resulting prompt--completion pairs using standard next-token prediction loss.

\subsection{Preliminaries}\label{sec:method:prelim}
We denote the calibration set by \(\mathcal{D}_{\mathrm{cal}} = \{(q^{(i)}, y^{(i)})\}_{i=1}^{N_{\mathrm{cal}}}\), where each example \( (q^{(i)}, y^{(i)})\) contains both a prompt and its answer.
We further denote the training-prompt set by \(\mathcal{D}_{\mathrm{train}} = \{q^{(i)}\}_{i=1}^{N_{\mathrm{train}}}\), where each example contains only the prompt.
After self-generation, we obtain a self-generated corpus \(\mathcal{D}_{\mathrm{corpus}} = \{(q^{(i)}, \hat{y}^{(i)})\}_{i=1}^{N_{\mathrm{train}}}\), where \(\hat{y}^{(i)}\) is the model-generated output for prompt \(q^{(i)}\).
%
Finally, let \(f_\theta\) denote a LLM with \(L\) layers, and let \(\mathcal{L} \subseteq \{1,\ldots,L\}\) denote the set of target layers on which our method operates. 
Motivated by prior work showing that intermediate representations provide useful training signals~\citep{romero2015fitnets,sun2019patient}, we by default apply projection to both K and V at a middle and a late layer, ($\mathcal{L} = {L, \lfloor L/2 \rfloor}$), unless otherwise noted.
%
%

\paragraph{Distillation paradigm comparison.} 
Let \(f_T\) denote an external teacher model, \(f_{\theta_{\mathrm{old}}}\) denote the student model before the distillation update, and \(f_{\mathrm{off}}\) denote an offline model induced by a teacher, or a fixed offline dataset. 
Let \(p\) denote the corresponding next-token probability distribution, and let \( z= T(q,y) = (z_1,\ldots,z_T)\) be tokenized sequence.
Many distillation methods can be written as matching a teacher distribution along trajectories sampled from a rollout policy \(f_{\mathrm{roll}}\):
\begin{equation}
\min_{\theta}
\mathbb{E}_{q\sim\mathcal{D},\, y\sim f_{\mathrm{roll}}(\cdot\mid q)}
\sum_{t=1}^{T}
\mathrm{KL}
\left(
p_T(\cdot \mid z_{<t})
\;\|\;
p_{\theta}(\cdot \mid z_{<t})
\right) 
\qquad
z={T}(q,y).
\end{equation}
In off-policy distillation, the prefixes \(z_{<t}\) come from an offline source, i.e., \(f_{\mathrm{roll}}=f_{\mathrm{off}}\), where \(f_{\mathrm{off}}\in\{f_T, \mathcal{D}_{\mathrm{offline}}\}\).
In on-policy distillation, the prefixes are sampled from the student policy itself, e.g., \(f_{\mathrm{roll}}=f_{\theta_{\mathrm{old}}}\).
Both paradigms require an external teacher \(f_T\) to provide token-level supervision.

Self-distillation, on the other hand, removes the external teacher and trains the model on its own generated outputs.
To improve data quality, existing methods introduce an external scoring or filtering signal \(S(q,y)\), leading to a weighted objective:
\begin{equation}
\min_{\theta}
-
\mathbb{E}_{q\sim\mathcal{D},\, y\sim f_{\theta_{\mathrm{old}}}(\cdot \mid q)}
\left[
S(q,y)\sum_{t }
    \log p_{\theta}
    \left(
        z_{t}
        \mid
        z_{<t}
    \right)
\right]
\qquad
z={T}(q,y).
\end{equation}
Thus, although the outputs are self-generated, the learning signal still largely depends on an external verifier, reward, or filtering mechanism.
%

\paragraph{Self-policy distillation (\ours).}
In contrast, \ours does not train directly on raw outputs from the original
\(f_{\theta_{\mathrm{old}}}\).
Instead, it first induces a self-policy through an internal transformation:
\begin{equation}
f_{\theta_{\mathrm{old}}}^{\mathrm{hook}}
=
\mathcal{T}^{\mathrm{hook}}_{\theta_{\mathrm{old}}}
\left(
f_{\theta_{\mathrm{old}}}
\right),
\end{equation}
where \(\mathcal{T}^{\mathrm{hook}}_{\theta_{\mathrm{old}}}\) is implemented by projection hooks derived from a capability-specific subspace. 
We then sample training outputs from this internally transformed policy:
\begin{equation}
\hat{y} \sim f_{\theta_{\mathrm{old}}}^{\mathrm{hook}}(\cdot \mid q).
\end{equation}
Finally, the original model is trained to absorb this capability-steered behavior:
\begin{equation}
\min_{\theta}
-
\mathbb{E}_{q\sim\mathcal{D},\, \hat{y}\sim f_{\theta_{\mathrm{old}}}^{\mathrm{hook}}(\cdot \mid q)}
\left[
\sum_{t }
    \log p_{\theta}
    \left(
        z_{t}
        \mid
        z_{<t}
    \right)
\right] \qquad
z={T}(q,\hat{y}).
\end{equation}
Equivalently, we have \(f_{\theta_{\mathrm{old}}}
\;\xrightarrow{\;\mathcal{T}^{\mathrm{hook}}_{\theta_{\mathrm{old}}}\;}\;
f_{\theta_{\mathrm{old}}}^{\mathrm{hook}}
\;\xrightarrow{\;\mathrm{generate}\;}\;
\hat{y}
\;\xrightarrow{\;\mathrm{distill}\;}\;
f_\theta .\)
Therefore, \ours can be viewed as distilling an internally selected version of the model's own policy back into the original model.
In the following sections, we describe the main procedure for extracting this capability-specific subspace and using it to induce the hooked self-policy.
\algref{alg:spd} provides a detailed overview of the full procedure.
%
%

\subsection{Capability Subspace Extraction}
\label{sec:method:subspace}
The goal of this phase is to select a low-dimensional subspace in the KV activation space that is most relevant to the target capability. 
Instead of using the standard next-token loss over the entire token sequence, we define a \textit{correctness-aligned loss} that only select token positions directly tied to task success. 
This design helps the extracted subspace focus on capability-relevant directions, while reducing the influence of stylistic tokens, formatting artifacts, and other task-irrelevant noise.   
We then collect token-level gradients from each calibration example and concatenate them across the full calibration set to form gradient matrices for K and V.
By applying SVD to these matrices, we retain the dominant directions in the KV dimensions and construct fixed projection matrices for later hook-based self-generation.

\paragraph{Correctness-aligned gradient collection.} 
After tokenizing each calibration example \((q^{(i)}, y^{(i)})\), we denote the resulting token sequence by \( T(q^{(i)}, y^{(i)})
= z^{(i)}=
(z^{(i)}_1,\ldots,z^{(i)}_{T_i}),\) where \(T_i\) is the token sequence length. 
%
Let \(\mathcal{S}^{(i)} \subseteq \{1,\ldots,T_i\}\) denote the set of correctness-defining target token positions.
These are token positions whose prediction is directly tied to task success. 
The details of how we select these token positions are provided in \appref{app: correctness-defining spans}.
Formally, we define the correctness-aligned loss as
\begin{equation}
    \mathcal{L}_{\mathrm{align}}(q^{(i)}, y^{(i)})
    =
    - \frac{1}{\lvert\mathcal{S}^{(i)}\rvert}
    \sum_{t \in \mathcal{S}^{(i)}}
    \log p_{\theta_{\mathrm{old}}}
    \left(
        z^{(i)}_{t}
        \mid
        z^{(i)}_{<t}
    \right).
    \label{eq:loss-aligned}
\end{equation}
For positions \(t \notin \mathcal{S}^{(i)}\), the labels are masked and do not contribute to the loss. Thus, the gradient signal is concentrated on tokens that are directly related to correctness.

For each calibration example \((q^{(i)}, y^{(i)})\), we perform a single forward--backward pass on the frozen student model. 
At each target layer \(\ell \in \mathcal{L}\), we compute the gradient of the aligned loss with respect to the key and value activation vectors at every token position:
\begin{equation}
    g_{K,t}^{(\ell,i)}
    =
    \frac{\partial \mathcal{L}_{\mathrm{align}}(q^{(i)}, y^{(i)})}
    {\partial K_t^{(\ell,i)}}
    \in \mathbb{R}^{d_k},
    \qquad
    g_{V,t}^{(\ell,i)}
    =
    \frac{\partial \mathcal{L}_{\mathrm{align}}(q^{(i)}, y^{(i)})}
    {\partial V_t^{(\ell,i)}}
    \in \mathbb{R}^{d_v},
\end{equation}
%
where $t=1,\ldots,T_i$; \(g_{K,t}^{(\ell,i)}\), \(g_{V,t}^{(\ell,i)}\) denote the key and value gradient vectors of token \(t\) at layer \(\ell\); \(d_k\) and \(d_v\) are the  KV dimensions. 
Although gradients are defined for all token positions, only positions connected to the correctness-aligned loss receive task-relevant gradient signals.

We then collect these token-level gradients from each calibration example and concatenate them across the full calibration set to form gradient matrices for K and V:
\begin{equation}
    G_K^{(\ell)}
    =
    \mathrm{Concat}_{i,t}
    \left(
    g_{K,t}^{(\ell,i)}
    \right)
    \in \mathbb{R}^{M \times d_k},
    \qquad
    G_V^{(\ell)}
    =
    \mathrm{Concat}_{i,t}
    \left(
    g_{V,t}^{(\ell,i)}
    \right)
    \in \mathbb{R}^{M \times d_v},
\end{equation}
where the concatenation is taken over the token positions \((i,t) \in \{(i,t): i=1,\ldots,N_{\mathrm{cal}},\; t=1,\ldots,T_i\}\), and
\(
    M =  \sum_{i=1}^{N_{\mathrm{cal}}} T_i
\). 
Therefore, each row of \(G_K^{(\ell)}\) or \(G_V^{(\ell)}\) corresponds to one token-level gradient direction in the KV feature space.

\paragraph{SVD and projection matrix.} 
For each target layer \(\ell \in \mathcal{L}\), we apply SVD to the collected gradient matrices.
For the key (K) activations, we write 
\begin{equation}
    G_K^{(\ell)}
    =
    U_K^{(\ell)} \Sigma_K^{(\ell)}
    \big(V_K^{(\ell)}\big)^\top .
\end{equation}
The right-singular vectors span directions in the K feature dimension. 
We retain the top-\(r\) right-singular vectors, \(  V_{K,r}^{(\ell)} \in \mathbb{R}^{d_k \times r},\) which correspond to the dominant gradient directions associated with correctness-defining tokens. 
The rank-\(r\) orthogonal projection matrix is then defined as 
\begin{equation}
    P_K^{(\ell)}
    =
    V_{K,r}^{(\ell)}
    \big(V_{K,r}^{(\ell)}\big)^\top
    \in \mathbb{R}^{d_k \times d_k}.
    \label{eq:projection-k}
\end{equation}
Similarly, for the value activations, we obtain
\begin{equation}
    P_V^{(\ell)}
    =
    V_{V,r}^{(\ell)}
    \big(V_{V,r}^{(\ell)}\big)^\top
    \in \mathbb{R}^{d_v \times d_v}.
    \label{eq:projection-v}
\end{equation}
Intuitively, the top singular directions capture the KV dimensions most sensitive to the correctness-aligned loss. 
Directions associated with smaller singular values are treated as lower-energy variation and are less likely to represent stable capability-relevant information. 
Therefore, the projection matrices \(P_K^{(\ell)}\) and \(P_V^{(\ell)}\) preserve the selected capability subspace while suppressing noisier directions. 
These matrices are computed once per run and kept fixed during subsequent self-generation.

\subsection{Capability-Selective Distillation}
\label{sec:method:hook}

After obtaining the projection matrices \(P_K^{(\ell)}\) and \(P_V^{(\ell)}\) from capability subspace extraction (\secref{sec:method:subspace}), we directly apply them as \emph{projection hooks} during self-generation. 
Specifically, at each target layer, the hook projects the KV activations onto the selected capability subspace. 
The goal is to produce self-generated outputs that are biased toward the target capability while keeping the model parameters unchanged. 
Finally, we fine-tune the original model on the resulting self-generated corpus.

\paragraph{Projection hooks.}
During autoregressive generation, we insert projection hooks at each target layer \(\ell \in \mathcal{L}\). 
Specifically, Projection hooks are applied to the K and V activations:
\begin{equation}
    \widetilde{K}^{(\ell)}
    =
    K^{(\ell)} P_K^{(\ell)},
    \qquad
    \widetilde{V}^{(\ell)}
    =
    V^{(\ell)} P_V^{(\ell)} .
    \label{eq:hook}
\end{equation}
The subsequent attention computation uses \(\widetilde{K}^{(\ell)}\) and \(\widetilde{V}^{(\ell)}\) instead of the original activations \(K^{(\ell)}\) and \(V^{(\ell)}\). 
Importantly, the model parameters are not updated during this step. 
The hooks only filter the intermediate KV activations through the selected capability subspace, thereby selectively preserving target capability signals while suppressing task-irrelevant variation.

\paragraph{Self-generation under projection.}
Let \(\mathcal{D}_{\mathrm{train}}=\{q^{(i)}\}_{i=1}^{N_{\mathrm{train}}}\) denote the  training-prompt set \textit{without ground truths}. 
With the projection hooks active, we sample one completion for each prompt:
\begin{equation}
   \hat{y}^{(i)} \sim f_{\theta_{\mathrm{old}}}^{\mathrm{hook}}(\cdot \mid q^{(i)}),
\quad i = 1,\ldots,N_{\mathrm{train}}  .
\end{equation}
This gives the self-generated corpus \(  \mathcal{D}_{\mathrm{corpus}}
    =
    \{(q^{(i)}, \hat{y}^{(i)})\}_{i=1}^{N_{\mathrm{train}}}.\)
Each \(\hat{y}^{(i)}\) remains the student's own raw generation, possibly incorrect, but it is generated under KV activations projected onto the capability subspace. 
The projection hooks only shape the self-generated training distribution.
During fine-tuning and final evaluation, all hooks are removed.

\paragraph{Supervised fine-tuning.} 
The last step applies LoRA~\citep{hu2022lora} to fine-tune the model on the self-generated corpus. 
Given a self-generated pair \((q^{(i)}, \hat{y}^{(i)}) \in \mathcal{D}_{\mathrm{corpus}}\), we concatenate and tokenize the prompt and completion as \( z^{(i)}=T(q^{(i)},\hat{y}^{(i)}).\) 
We then optimize the standard next-token prediction loss:
\begin{equation}
    \min_{\Delta\theta_{\mathrm{LoRA}}}
    -
    \sum_{(q^{(i)}, \hat{y}^{(i)}) \in \mathcal{D}_{\mathrm{corpus}}}
    \sum_{t=2}^{|z^{(i)}|}
    \log p_{\theta_{\mathrm{old}}+\Delta\theta_{\mathrm{LoRA}}}
    \left(
        z^{(i)}_{t}
        \mid
        z^{(i)}_{<t}
    \right).
\end{equation}
Here, \(\Delta\theta_{\mathrm{LoRA}}\) denotes the trainable LoRA parameters.
Importantly, the projection hooks in \eqnref{eq:hook} are used only during self-generation and are removed during fine-tuning. 
Thus, \ours remains a standard self-distillation procedure at training time, optimizing the student with ordinary next-token prediction on its own generated outputs. 
%


\section{Experiments}
\label{sec:experiments}
\subsection{Experimental Setup}

\begin{table*}[t]
\centering
\small
\caption{
    Domain-specific and in-domain generalization results under different backbone models.
}
\vspace{-0.1in}
\setlength{\tabcolsep}{12pt}
\resizebox{\textwidth}{!}{%
\begin{tabular}{llcccccc}
\toprule
& & \multicolumn{2}{c}{Code} & \multicolumn{2}{c}{Math} & \multicolumn{2}{c}{QA} \\
\cmidrule(lr){3-4}\cmidrule(lr){5-6}\cmidrule(lr){7-8}
Backbone & Method
& MBPP \(\uparrow\)
& CodeAlpaca \(\downarrow\)
& GSM8K \(\uparrow\)
& SVAMP \(\uparrow\)
& MMLU \(\uparrow\)
& BBH  \(\uparrow\) \\
\midrule

\multirow{4}{*}{Qwen2.5-0.5B-Instruct}
& Base model
& \hlr{\(\res{17.0}{1.2}\%\)}
& \hlg{\(\res{0.683}{0.000}\)}
& \hlr{\(\res{11.0}{0.0}\%\)}
& \hlg{\(\res{16.0}{0.0}\%\)}
& \hlr{\(\res{46.0}{0.6}\%\)}
& \hlg{\(\res{32.7}{0.0}\%\)} \\

& PSR
& \hlr{\(\mathbf{\res{28.7}{2.9}\%}\)}
& \hlg{\(\res{0.684}{0.001}\)}
& \hlr{\(\underline{\res{17.0}{2.0}\%}\)}
& \hlg{\(\underline{\res{19.3}{4.4}\%}\)}
& \hlr{\(\res{42.7}{0.6}\%\)}
& \hlg{\(\res{33.1}{1.2}\%\)} \\

& SSD
& \hlr{\(\res{18.3}{4.5}\%\)}
& \hlg{\(\underline{\res{0.683}{0.001}}\)}
& \hlr{\(\res{12.3}{2.9}\%\)}
& \hlg{\(\res{16.7}{0.6}\%\)}
& \hlr{\(\underline{\res{48.0}{0.0}\%}\)}
& \hlg{\(\underline{\res{35.7}{0.6}\%}\)} \\

& \ours
& \hlr{\(\underline{\res{27.0}{2.0}\%}\)}
& \hlg{\(\mathbf{\res{0.678}{0.001}}\)}
& \hlr{\(\mathbf{\res{23.0}{2.6}\%}\)}
& \hlg{\(\mathbf{\res{32.3}{1.2}\%}\)}
& \hlr{\(\mathbf{\res{49.7}{0.6}\%}\)}
& \hlg{\(\mathbf{\res{38.1}{0.6}\%}\)} \\
\midrule

\multirow{4}{*}{Qwen2.5-7B-Instruct}
& Base model
& \hlr{\(\underline{\res{58.5}{0.2}\%}\)}
& \hlg{\(\res{0.653}{0.000}\)}
& \hlr{\(\underline{\res{49.0}{0.0}\%}\)}
& \hlg{\(\underline{\res{72.0}{0.0}\%}\)}
& \hlr{\(\res{65.0}{0.0}\%\)}
& \hlg{\(\res{47.3}{0.0}\%\)} \\

& PSR
& \hlr{\(\res{38.7}{2.0}\%\)}
& \hlg{\(\underline{\res{0.606}{0.002}}\)}
& \hlr{\(\res{47.3}{2.3}\%\)}
& \hlg{\(\res{69.7}{1.7}\%\)}
& \hlr{\(\mathbf{\res{71.7}{3.7}\%}\)}
& \hlg{\(\res{48.3}{2.7}\%\)} \\

& SSD
& \hlr{\(\res{57.1}{0.6}\%\)}
& \hlg{\(\res{0.623}{0.004}\)}
& \hlr{\(\res{28.0}{1.2}\%\)}
& \hlg{\(\res{68.0}{1.7}\%\)}
& \hlr{\(\res{67.7}{4.5}\%\)}
& \hlg{\(\underline{\res{45.7}{1.2}\%}\)} \\

& \ours
& \hlr{\(\mathbf{\res{68.8}{0.2}\%}\)}
& \hlg{\(\mathbf{\res{0.596}{0.001}}\)}
& \hlr{\(\mathbf{\res{50.0}{1.7}\%}\)}
& \hlg{\(\mathbf{\res{80.0}{0.6}\%}\)}
& \hlr{\(\underline{\res{70.7}{1.5}\%}\)}
& \hlg{\(\mathbf{\res{50.3}{1.3}\%}\)} \\
\midrule

\multirow{4}{*}{Qwen2.5-14B-Instruct}
& Base model
& \hlr{\(\underline{\res{70.8}{0.0}\%}\)}
& \hlg{\(\res{0.739}{0.002}\)}
& \hlr{\(\underline{\res{57.0}{0.0}\%}\)}
& \hlg{\(\res{77.0}{0.0}\%\)}
& \hlr{\(\res{72.5}{0.0}\%\)}
& \hlg{\(\res{54.0}{0.0}\%\)} \\

& PSR
& \hlr{\(\res{67.6}{2.0}\%\)}
& \hlg{\(\res{0.694}{0.004}\)}
& \hlr{\(\res{55.0}{3.0}\%\)}
& \hlg{\(\underline{\res{81.0}{1.0}\%}\)}
& \hlr{\(\res{72.0}{1.0}\%\)}
& \hlg{\(\mathbf{\res{55.3}{1.1}\%}\)} \\

& SSD
& \hlr{\(\res{68.5}{1.5}\%\)}
& \hlg{\(\underline{\res{0.689}{0.001}}\)}
& \hlr{\(\res{46.0}{2.6}\%\)}
& \hlg{\(\res{78.0}{2.0}\%\)}
& \hlr{\(\underline{\res{73.5}{1.5}\%}\)}
& \hlg{\(\res{54.0}{1.6}\%\)} \\

& \ours
& \hlr{\(\mathbf{\res{75.4}{1.5}\%}\)}
& \hlg{\(\mathbf{\res{0.687}{0.001}}\)}
& \hlr{\(\mathbf{\res{60.0}{1.7}\%}\)}
& \hlg{\(\mathbf{\res{84.0}{1.5}\%}\)}
& \hlr{\(\mathbf{\res{74.5}{0.9}\%}\)}
& \hlg{\(\underline{\res{54.7}{0.7}\%}\)} \\
\midrule

\multirow{4}{*}{Qwen3-4B-Instruct}
& Base model
& \hlr{\(\underline{\res{65.4}{0.0}\%}\)}
& \hlg{\(\res{0.809}{0.000}\)}
& \hlr{\(\res{30.0}{0.0}\%\)}
& \hlg{\(\underline{\res{18.0}{0.0}\%}\)}
& \hlr{\(\res{63.0}{0.0}\%\)}
& \hlg{\(\underline{\res{49.0}{0.0}\%}\)} \\

& PSR
& \hlr{\(\res{63.1}{0.6}\%\)}
& \hlg{\(\underline{\res{0.754}{0.002}}\)}
& \hlr{\(\res{29.0}{0.0}\%\)}
& \hlg{\(\res{12.7}{2.3}\%\)}
& \hlr{\(\res{61.5}{0.2}\%\)}
& \hlg{\(\res{48.7}{0.3}\%\)} \\

& SSD
& \hlr{\(\res{63.1}{2.9}\%\)}
& \hlg{\(\res{0.756}{0.004}\)}
& \hlr{\(\underline{\res{30.7}{0.6}\%}\)}
& \hlg{\(\res{16.7}{1.2}\%\)}
& \hlr{\(\underline{\res{62.7}{0.6}\%}\)}
& \hlg{\(\res{48.0}{0.2}\%\)} \\

& \ours
& \hlr{\(\mathbf{\res{70.8}{1.6}\%}\)}
& \hlg{\(\mathbf{\res{0.727}{0.001}}\)}
& \hlr{\(\mathbf{\res{33.0}{1.0}\%}\)}
& \hlg{\(\mathbf{\res{23.0}{1.5}\%}\)}
& \hlr{\(\mathbf{\res{65.8}{0.3}\%}\)}
& \hlg{\(\mathbf{\res{49.3}{0.2}\%}\)} \\
\midrule

\multirow{4}{*}{Llama-3.1-8B-Instruct}
& Base model
& \hlr{\(\res{28.4}{0.0}\%\)}
& \hlg{\(\res{0.751}{0.000}\)}
& \hlr{\(\underline{\res{39.0}{0.6}\%}\)}
& \hlg{\(\underline{\res{55.0}{0.0}\%}\)}
& \hlr{\(\res{60.0}{0.0}\%\)}
& \hlg{\(\mathbf{\res{59.0}{0.0}\%}\)} \\

& PSR
& \hlr{\(\underline{\res{38.0}{0.0}\%}\)}
& \hlg{\(\res{0.786}{0.001}\)}
& \hlr{\(\res{26.0}{0.0}\%\)}
& \hlg{\(\res{49.0}{1.2}\%\)}
& \hlr{\(\underline{\res{62.0}{0.0}\%}\)}
& \hlg{\(\res{56.7}{1.6}\%\)} \\

& SSD
& \hlr{\(\res{31.4}{1.2}\%\)}
& \hlg{\(\underline{\res{0.729}{0.002}}\)}
& \hlr{\(\res{32.0}{1.0}\%\)}
& \hlg{\(\res{47.7}{3.5}\%\)}
& \hlr{\(\res{58.3}{1.5}\%\)}
& \hlg{\(\res{56.7}{0.0}\%\)} \\

& \ours
& \hlr{\(\mathbf{\res{41.4}{2.0}\%}\)}
& \hlg{\(\mathbf{\res{0.692}{0.000}}\)}
& \hlr{\(\mathbf{\res{40.0}{2.0}\%}\)}
& \hlg{\(\mathbf{\res{56.7}{1.0}\%}\)}
& \hlr{\(\mathbf{\res{62.5}{0.6}\%}\)}
& \hlg{\(\underline{\res{57.3}{0.5}\%}\)} \\

\bottomrule
\end{tabular}}
\label{tab:main_results}
\vspace{-0.15in}
\end{table*}

\paragraph{Datasets.}
We evaluate \ours across three capability domains:
(1) code generation, including MBPP~\citep{Austin2021ProgramSW} and CodeAlpaca-20k~\citep{codealpaca};
(2) mathematical reasoning, including GSM8K~\citep{hosseini2024v} and SVAMP~\citep{Patel2021AreNM};
and (3) question answering (QA), including MMLU~\citep{Hendrycks2020MeasuringMM} and BBH~\citep{Suzgun2022ChallengingBT}.

For each capability domain, we use one dataset as the source benchmark for training prompts, calibration, and \rosebox{source-benchmark evaluation}. 
We additionally evaluate on a second held-out dataset from the same capability domain and and on datasets from other capability domains to measure \purplebox{in-domain generalization} and \orangebox{out-of-domain generalization}, respectively.
This setup allows us to test whether \ours improves the source benchmark itself, and whether the learned gains transfer to another dataset in the same capability domain rather than merely overfitting to dataset-specific patterns.

\paragraph{Baselines and LLM backbones.}
We compare \ours against three baselines.
(1) \textit{Base model}: the original pretrained student that measures the model's pre-existing capability on each task.
(2) \textit{Plain Self-Retraining (PSR)}: the student fine-tuned on its own raw outputs generated by the base model, without projection hooks or additional filtering. 
This baseline isolates the effect of simply retraining on self-generated data.
And (3) \textit{Simple Self-Distillation (SSD)}: the self-distillation baseline adapted from~\citet{zhang2026ssd}, where the model is trained on its own self-generated data produced with truncation-based decoding. 
We instantiate each methods with five open-source LLM backbones spanning two models families and various sizes: Qwen2.5-0.5B-Instruct, Qwen2.5-7B-Instruct, Qwen2.5-14B-Instruct~\citep{yang2025qwen25technicalreport},  Qwen3-4B-Instruct~\citep{yang2025qwen3technicalreport} and Llama-3.1-8B-Instruct~\citep{Dubey2024TheL3}. 
This range lets us verify that \ours's benefit do not depend on a particular model family, scale. 
(Hyperparameter and implementation details are provided in \appref{subsec:implementation_details}.). 

\paragraph{Evaluation metrics.}
%

We report \textsc{pass@1} on MBPP, exact-match accuracy on GSM8K and SVAMP, answer-letter accuracy on MMLU, and normalized exact-match on BBH, along with negative log-likelihood on CodeAlpaca.
Higher is better for all metrics except NLL, where lower is better.

\subsection{Main Results}\label{sec:exp:main}
\tabref{tab:main_results} summarizes the main domain-specific performance and in-domain generalization transfer results.  
Across code generation, mathematical reasoning, and multiple-choice QA, \ours consistently outperforms the base model, PSR, and SSD across five backbone models.
Among the five backbones, the largest average relative improvements of \ours reach 9.5\%, 9.2\%, and 10.7\% over the base model, PSR, and SSD, respectively.

Moreover, \ours demonstrates strong in-domain generalization.
It improves over the base model on  14 out of 15 in-domain transfer evaluations, whereas PSR improves on only 7 out of 15 and SSD improves on only 8 out of 15. 
This shows that the gains of \ours are not limited to the source benchmark, but transfer more reliably to held-out datasets within the same capability domain.

\subsection{Out-of-Domain Generalization}
\begin{wraptable}{r}{0.5\textwidth}
\vspace{-0.4in}
\centering
\small
\caption{Out-of-domain generalization from QA-calibrated self-policy distillation under Qwen2.5-0.5B-Instruct.}
\label{tab:out_of_domain}
\vspace{-0.1in}
\setlength{\tabcolsep}{4pt}
\resizebox{\linewidth}{!}{%
\begin{tabular}{lcccccc}
\toprule
Method
& \multicolumn{2}{c}{QA}
& \multicolumn{2}{c}{Math}
& \multicolumn{2}{c}{Code} \\
\cmidrule(lr){2-3}\cmidrule(lr){4-5}\cmidrule(lr){6-7}
& MMLU \(\uparrow\)
& BBH \(\uparrow\)
& GSM8K \(\uparrow\)
& SVAMP \(\uparrow\)
& MBPP \(\uparrow\)
& CodeAlpaca \(\downarrow\)\\
\midrule

Base model
& \hlr{\(\res{46.0}{0.6}\%\)}
& \hlg{\(\res{32.7}{0.0}\%\)}
& \hlb{\(\res{11.0}{0.0}\%\)}
& \hlb{\(\underline{\res{16.0}{0.0}}\%\)}
& \hlb{\(\underline{\res{17.0}{1.2}}\%\)}
& \hlb{\(\res{0.683}{0.000}\)} \\

SSD-QA
& \hlr{\(\underline{\res{48.0}{1.2}}\%\)}
& \hlg{\(\underline{\res{36.0}{0.6}}\%\)}
& \hlb{\(\underline{\res{19.0}{2.2}}\%\)}
& \hlb{\(\res{14.3}{2.0}\%\)}
& \hlb{\(\res{12.0}{1.2}\%\)}
& \hlb{\(\underline{\res{0.676}{0.008}}\)} \\

SPD-QA
& \hlr{\(\mathbf{\res{49.7}{0.6}}\%\)}
& \hlg{\(\mathbf{\res{38.1}{0.6}}\%\)}
& \hlb{\(\mathbf{\res{26.7}{3.2}}\%\)}
& \hlb{\(\mathbf{\res{19.3}{2.5}}\%\)}
& \hlb{\(\mathbf{\res{21.3}{0.6}}\%\)}
& \hlb{\(\mathbf{\res{0.675}{0.002}}\)} \\

\bottomrule
\end{tabular}}
\label{tab:ablation_result}
\vspace{-0.15in}
\end{wraptable}
\tabref{tab:out_of_domain} highlights the capability selectivity of our method.
Here, the subspace is extracted using only the QA domain, yet \textit{\ours-QA} improves not only on MMLU and BBH, but also on Math and Code benchmarks. 
This suggests that the capability filtered by \ours is not limited to surface answer format, but captures a more transferable reasoning signal that can benefit multiple domains. 
This pattern is also consistent with the broader view that structured reasoning learned from one domain can transfer beyond that domain. 
Compared with SSD-QA, \ours-QA delivers consistently stronger gains, showing that capability-selective projection produces self-generated data with a more useful and selective focus. 
In this sense, \ours enables \emph{capability selectivity}: the calibration domain determines extracted capability subspace, allowing us to steer self-generation toward a target capability and enable more effective transfer across tasks.


\section{Ablation and Analysis}
\subsection{Self-Generated Data with Projection Hooks}
\begin{figure}[t]  
    \centering
    \includegraphics[width=1\linewidth]{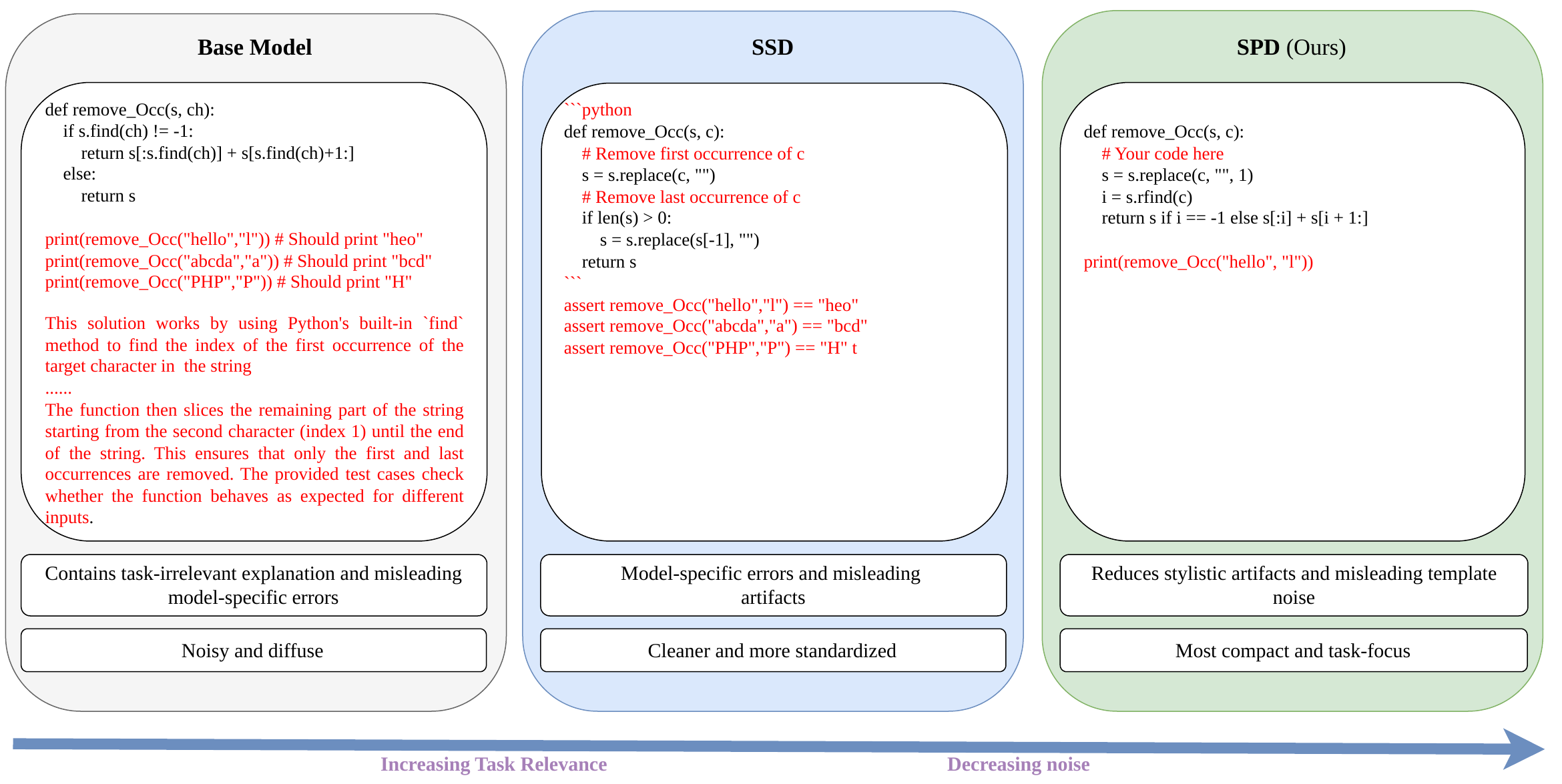}  
    \vspace{-0.2in}
    \caption{Self-generated data output comparisons under Qwen2.5-0.5B-Instruct. Black text denotes task-relevant content, while red text highlights answer-irrelevant or low-value content.}
    \label{fig:case_study}
    \vspace{-0.15in}
\end{figure}
We analyze the role of projection hooks from three complementary views: the raw output patterns they produce, the quality of the resulting self-generated data before training, and the downstream performance after fine-tuning on that data.
%
%

\paragraph{Qualitative example of self-generated data.}
\figref{fig:case_study} illustrates different ways in which self-generated data can be shaped. 
The base model output is verbose and unfocused: it mixes the function implementation with extra \texttt{print} statements and a long explanation, so the core task signal is diluted by non-essential text. 
The SSD truncated output is cleaner and more standardized, following a ``function + comments + assertions'' template. 
However, it still contains incorrect program logic and misleading verification cues, so its improvement is mainly in surface form rather than in implementing the required behavior. 
In contrast, the \ours self-generated output is the most compact and task-focused: it removes most explanatory text and template filler, and concentrates generation on the core code fragment. 
This highlights the key advantage of our method: rather than only regularizing raw outputs after generation, it steers the generation process itself toward capability-relevant content. 
As a result, the produced self-generated data is more targeted, less noisy, and less affected by stylistic or formatting artifacts, making it more suitable for subsequent fine-tuning.

\paragraph{Quantitative comparisons of self-generated data.}
\begin{wraptable}{r}{0.6\textwidth}
\vspace{-0.2in}
\centering
\small
\caption{Quality of self-generated data before self-distillation training under Qwen2.5-0.5B-Instruct.}
\setlength{\tabcolsep}{4pt}
\resizebox{\linewidth}{!}{%
\begin{tabular}{lcccccc}
\toprule
Method
& \multicolumn{2}{c}{Code}
& \multicolumn{2}{c}{Math}
& \multicolumn{2}{c}{QA} \\
\cmidrule(lr){2-3}\cmidrule(lr){4-5}\cmidrule(lr){6-7}
& MBPP \(\uparrow\)
& CodeAlpaca \(\downarrow\)
& GSM8K \(\uparrow\)
& SVAMP \(\uparrow\)
& MMLU \(\uparrow\)
& BBH  \(\uparrow\) \\
\midrule

Base-Generated Data
& \hlr{\(\mathbf{\res{17.0}{1.2}}\%\)}
& \hlg{\(\mathbf{\res{0.683}{0.000}}\)}
& \hlr{\(\underline{\res{11.0}{0.0}}\%\)}
& \hlg{\(\mathbf{\res{16.0}{0.0}}\%\)}
& \hlr{\(\res{46.0}{0.6}\%\)}
& \hlg{\(\underline{\res{32.7}{0.0}}\%\)} \\

SSD-Generated Data
& \hlr{\(\res{15.0}{0.2}\%\)}
& \hlg{\(\mathbf{\res{0.683}{0.000}}\)}
& \hlr{\(\res{10.7}{1.2}\%\)}
& \hlg{\({\res{14.0}{0.0}}\%\)}
& \hlr{\(\underline{\res{46.5}{0.6}}\%\)}
& \hlg{\(\res{28.7}{1.2}\%\)} \\

\ours-Generated Data
& \hlr{\(\underline{\res{16.3}{0.0}}\%\)}
& \hlg{\(\res{0.705}{0.002}\)}
& \hlr{\(\mathbf{\res{18.3}{0.6}}\%\)}
& \hlg{\(\underline{\res{14.7}{1.5}}\%\)}
& \hlr{\(\mathbf{\res{47.0}{1.0}}\%\)}
& \hlg{\(\mathbf{\res{35.3}{0.0}}\%\)} \\

\bottomrule
\end{tabular}}
\label{tab:self-generated_data_quality}
\vspace{-0.15in}
\end{wraptable}
\tabref{tab:self-generated_data_quality} compares the quantitative quality of self-generated data produced by three generation strategies before any self-distillation training: \textit{Base-Generated Data}, \textit{SSD-Generated Data}, and \textit{SPD-Generated Data}. 
Compared with \textit{SSD-Generated Data}, \textit{SPD-Generated Data} achieves clear improvements on MBPP (+1.3\% points), GSM8K (+7.6\%), and BBH (+6.6\%), while with smaller gains on SVAMP (+0.7\%) and MMLU (+0.5\% ). 
%

Compared with \textit{Base-Generated Data}, \textit{SPD-Generated Data} also improves GSM8K (+7.3\%), MMLU (+1.0\%), and BBH (+2.6\%), indicating that capability-selective projection can improve the usefulness of self-generated data even before fine-tuning. 
%
Overall, these results suggest that projection hooks act as an internal capability filter, improving the usefulness of self-generated data even before fine-tuning and producing more targeted, higher-value supervision for the subsequent self-distillation stage.

\newcommand{\methodcell}[1]{\makebox[1.45cm][l]{#1}}

\paragraph{Analysis on fine-tuning performance.}
\begin{wraptable}{r}{0.5\textwidth}
\vspace{0in}
\centering
\small
\caption{Fine-tuning results on self-generated data from different methods  under Qwen2.5-0.5B-Instruct. }
\vspace{-0.15in}
\setlength{\tabcolsep}{4pt}
\resizebox{\linewidth}{!}{%
\begin{tabular}{lcccccc}
\toprule
Method
& \multicolumn{2}{c}{Code}
& \multicolumn{2}{c}{Math}
& \multicolumn{2}{c}{QA} \\
\cmidrule(lr){2-3}\cmidrule(lr){4-5}\cmidrule(lr){6-7}
& MBPP \(\uparrow\)
& CodeAlpaca \(\downarrow\)
& GSM8K \(\uparrow\)
& SVAMP \(\uparrow\)
& MMLU \(\uparrow\)
& BBH \(\uparrow\)\\
\midrule

PSR
& \hlr{\(\mathbf{\res{28.7}{2.9}}\%\)}
& \hlg{\(\res{0.684}{0.001}\)}
& \hlr{\(\underline{\res{17.0}{2.0}}\%\)}
& \hlg{\(\underline{\res{19.3}{4.4}}\%\)}
& \hlr{\(\res{42.7}{0.6}\%\)}
& \hlg{\(\res{33.1}{1.2}\%\)} \\

SSD
& \hlr{\(\res{18.3}{4.5}\%\)}
& \hlg{\(\underline{\res{0.683}{0.001}}\)}
& \hlr{\(\res{12.3}{2.9}\%\)}
& \hlg{\(\res{16.7}{0.6}\%\)}
& \hlr{\(\underline{\res{48.0}{0.0}}\%\)}
& \hlg{\(\underline{\res{35.7}{0.6}}\%\)} \\

\ours
& \hlr{\(\underline{\res{27.0}{2.0}}\%\)}
& \hlg{\(\mathbf{\res{0.678}{0.001}}\)}
& \hlr{\(\mathbf{\res{23.0}{2.6}}\%\)}
& \hlg{\(\mathbf{\res{32.3}{1.2}}\%\)}
& \hlr{\(\mathbf{\res{49.7}{0.6}}\%\)}
& \hlg{\(\mathbf{\res{38.1}{0.6}}\%\)} \\

\bottomrule
\end{tabular}}
\label{tab:self_generated_data_fine_tune_result}
\vspace{-0.15in}
\end{wraptable}

\tabref{tab:self_generated_data_fine_tune_result} reports the next stage of the pipeline, where the model is fine-tuned on self-generated data from different methods.
%
In the code domain, PSR achieves strong MBPP performance (28.7\%) but limited CodeAlpaca transfer (0.684)
This suggests that unfocused self-generated data may overfit the source benchmark without generalizing well. 
SSD performs worse on MBPP (18.3\%) and shows almost no improvement on CodeAlpaca (0.683), indicating that truncated self-generated outputs alone do not provide reliable supervision. 
In contrast, \ours achieves a better balance between domain-specific performance and in-domain generalization, reaching 27.0\% on MBPP and the best CodeAlpaca result (0.678). 
Similar trends also appear in math and QA. 
%
Overall, these results suggest that self-generated data produced by \ours provides more effective supervision for fine-tuning, leading to a better balance between domain-specific performance and in-domain generalization.








%
\paragraph{Ablation on loss functions}
\begin{wraptable}{r}{0.5\textwidth}
\vspace{-0.17in}
\centering
\small
\caption{Ablation on loss functions for capability subspace extraction under Qwen2.5-0.5B-Instruct.}
\vspace{-0.05in}
\setlength{\tabcolsep}{2pt}
\resizebox{\linewidth}{!}{%
\begin{tabular}{lcccccc}
\toprule
Method
& \multicolumn{2}{c}{Code}
& \multicolumn{2}{c}{Math}
& \multicolumn{2}{c}{QA} \\
\cmidrule(lr){2-3}\cmidrule(lr){4-5}\cmidrule(lr){6-7}
& MBPP \(\uparrow\)
& CodeAlpaca \(\downarrow\)
& GSM8K \(\uparrow\)
& SVAMP \(\uparrow\)
& MMLU \(\uparrow\)
& BBH \(\uparrow\) \\
\midrule

Base model
& \hlr{\(\underline{\res{17.0}{1.2}}\%\)}
& \hlg{\(\res{0.683}{0.000}\)}
& \hlr{\(\res{11.0}{0.0}\%\)}
& \hlg{\(\res{16.0}{0.0}\%\)}
& \hlr{\(\res{46.0}{0.6}\%\)}
& \hlg{\(\res{32.7}{0.0}\%\)} \\

\makecell[l]{\ours\\~~w/ Full-sequence loss}
& \hlr{\(\res{12.9}{1.5}\%\)}
& \hlg{\(\underline{\res{0.681}{0.003}}\)}
& \hlr{\(\underline{\res{13.3}{1.7}}\%\)}
& \hlg{\(\underline{\res{23.7}{1.0}}\%\)}
& \hlr{\(\underline{\res{48.0}{0.0}}\%\)}
& \hlg{\(\underline{\res{35.7}{1.2}}\%\)} \\

\makecell[l]{\ours\\~~w/ Correctness-aligned loss}
& \hlr{\(\mathbf{\res{27.0}{2.0}}\%\)}
& \hlg{\(\mathbf{\res{0.678}{0.001}}\)}
& \hlr{\(\mathbf{\res{23.0}{2.6}}\%\)}
& \hlg{\(\mathbf{\res{32.3}{1.2}}\%\)}
& \hlr{\(\mathbf{\res{49.7}{0.6}}\%\)}
& \hlg{\(\mathbf{\res{38.1}{0.6}}\%\)} \\

\bottomrule
\end{tabular}}
\label{tab:ablation_result}
\vspace{-0.15in}
\end{wraptable}

\tabref{tab:ablation_result} compares two choices for extracting the capability subspace: \textit{Full-sequence loss} and our proposed \textit{Correctness-aligned loss}. 
\textit{Full-sequence loss} is the standard next-token loss computed over all tokens in the output sequence. 
In contrast, \textit{Correctness-aligned loss} is computed only on correctness-defining token positions, i.e., tokens whose prediction directly determines task success. 
%
This design concentrates the gradient signal on capability-relevant directions and reduces the influence of stylistic or task-irrelevant tokens.
We find that \textit{Correctness-aligned loss} consistently yields better results across all three domains. 
Compared with \textit{Full-sequence loss}, it improves MBPP from 12.9\% to 27.0\%, GSM8K from 13.3\% to 23.0\%, SVAMP from 23.7\% to 32.3\%, MMLU from 48.0\% to 49.7\%, and BBH from 35.7\% to 38.1\%. 
%
These results suggest that focusing on correctness-defining positions yields a more capability-focused subspace, producing more effective self-generated supervision for fine-tuning.

\subsection{Robust Analysis}

%
\begin{wrapfigure}{r}{0.4\textwidth}
 \vspace{-0.4in}
    \centering
    \includegraphics[width=\linewidth]{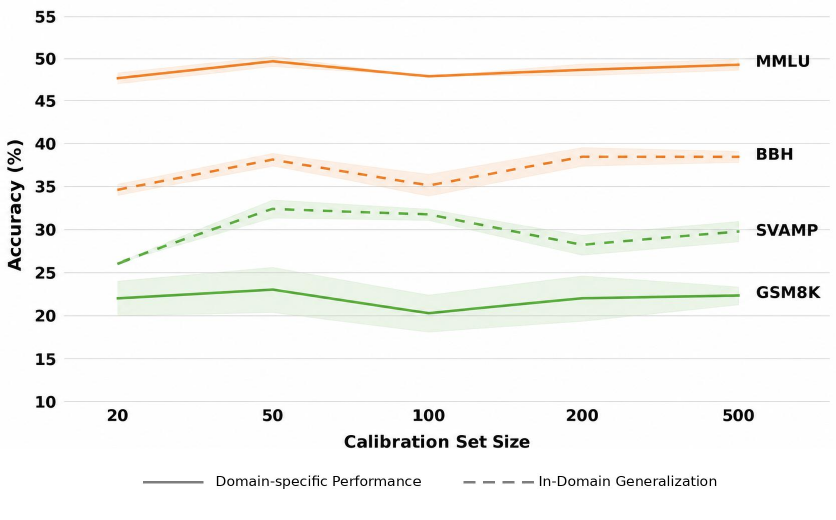}  
    \caption{Effect of calibration size  under Qwen2.5-0.5B-Instruct.}
    \label{fig:calibration}
\vspace{-0.1in}
\end{wrapfigure}
Calibration set size denotes the number of labeled examples used to extract the capability subspace. 
It determines how much task-specific signal is available for estimating the gradient directions that define the projection space. 
As shown in \figref{fig:calibration}, \ours does not require a large calibration set to remain effective, a modest number of labeled examples is enough to estimate useful capability subspaces.
Performance is stable across 20--500 calibration examples: in math, GSM8K stays around 20--23\% and SVAMP around 26--32\%; in QA, MMLU remains around 48--50\% and BBH around 34--38\%.
Overall, these results show that a small calibration set is sufficient for estimating stable and useful capability subspaces, highlighting the data efficiency of the calibration 
stage in \ours.
%


\section{Conclusion and Limitation}
In this paper, we argue that current self-distillation frameworks either rely on external signals or remain domain-specific and hard to generalize, while their raw self-generated outputs often contain task-irrelevant noise, all of which can reduce the effectiveness of self-distillation.
To address this issue, we propose Self-Policy Distillation (\ours), a capability-selective framework that extracts low-rank subspaces from correctness-aligned gradients, uses KV projection hooks to steer self-generation, and fine-tunes the model on the resulting self-generated outputs. 
Experiments across code generation, mathematical reasoning, and question answering show that \ours consistently improves over baselines and generalizes under both in-domain and out-of-domain transfer settings. 

\paragraph{Limitation.} 
Although \ours is evaluated across three capability domains and multiple backbones, broader validation on more diverse and high-stakes tasks would further strengthen its empirical robustness. 
We leave this direction for future work.
%


\bibliographystyle{iclr2027_conference}
\bibliography{iclr2027_conference}


\clearpage
\appendix

\section{Experimental Details}\label{app:experiment_details}

\subsection{Correctness-defining Spans}
\label{app: correctness-defining spans}
In this subsection, we describe how correctness-defining spans are selected for the three domains.
As shown in \tabref{tab:correctness-spans}, these spans consist of token positions whose predictions directly determine task success.
Concretely, for code generation, we use assertion-relevant spans because the assertions specify the functional behavior that the generated program must satisfy. Gradients from these tokens therefore emphasize the parts of the output that are directly linked to executable correctness. For mathematical reasoning, we use the final numeric answer span, since this span determines whether the solution is counted as correct under exact-match evaluation. For multiple-choice QA, we use the answer-letter position after \texttt{Answer:}, because the predicted option letter directly defines task success. These choices allow the calibration loss to focus on correctness-critical tokens rather than stylistic text, intermediate explanations, or other non-essential output tokens.

\begin{table}[h]
\centering
\caption{
    Correctness-defining spans used to construct the aligned loss in \eqnref{eq:loss-aligned}.
}
\small
\begin{tabular}{@{}lll@{}}
\toprule
Task & Correctness-defining target positions \(\mathcal{S}^{(i)}\) \\
\midrule
MBPP (code generation) & tokens in assertion-relevant spans \\
MMLU (QA) & the answer letter after \texttt{Answer:} \\
GSM8K (mathematical reasoning) & the final numeric answer span  \\
\bottomrule
\end{tabular}
\label{tab:correctness-spans}
\end{table}

\subsection{Dataset Descriptions}

\textbf{MBPP.}~\citep{Austin2021ProgramSW}is a code generation benchmark consisting of crowd-sourced Python programming problems. Each example contains a natural-language task description, a reference solution, and automated \texttt{assert} tests, making it suitable for evaluating functional correctness via execution.

\textbf{CodeAlpaca-20k.}~\citep{codealpaca} is an instruction-following code generation dataset containing 20K coding instructions. Each example includes an instruction, an optional input, and a model-generated output, making it suitable for evaluating general code instruction following beyond executable programming tests.  We evaluate CodeAlpaca using two metrics: negative log-likelihood (NLL), which measures the likelihood of reference outputs, and accuracy, which is used for accuracy-based summary  in ~\figref{fig:intro_teaser}.

\textbf{GSM8K.}~\citep{hosseini2024v} is a mathematical reasoning dataset containing grade-school-level word problems. Each example requires multi-step arithmetic reasoning and provides a final numeric answer, making it suitable for evaluating step-by-step mathematical problem solving.

\textbf{SVAMP.}~\citep{Patel2021AreNM}is a challenge set for arithmetic word problems. It is constructed by applying variations to existing math word problems, producing examples with different surface forms and reasoning patterns for evaluating mathematical transfer.

\textbf{MMLU.}~\citep{Hendrycks2020MeasuringMM} is a multiple-choice question answering benchmark covering 57 subjects across STEM, humanities, social sciences, and other domains. It evaluates broad world knowledge and problem-solving ability through answer-letter prediction.

\textbf{BBH.}~\citep{Suzgun2022ChallengingBT} is a collection of challenging BIG-Bench tasks selected because prior language model evaluations did not outperform average human-rater performance. It is commonly used to evaluate difficult reasoning and instruction-following capabilities.

\subsection{Hyperparameter}

\begin{table}[h]
\centering
\small
\caption{\texttt{\ours} hyperparameters.}
\label{tab:hyperparameters}

\begin{tabularx}{\linewidth}{lX}
\toprule
\textbf{Hyperparameter} & \textbf{Default} \\
\midrule

\texttt{model} &
\makecell[l]{
\texttt{Qwen/Qwen2.5-(0.5/7B/14B)-Instruct,} \\
\texttt{Qwen/Qwen3-4B-Instruct}\\
\texttt{Qwen/Qwen3.5-4B}\\
\texttt{LLama/Llama-3.1-8B-Instruct}
} \\

\texttt{task} & \texttt{math/code/QA} \\
\texttt{n\_train} & 7473/464/2000 \\
\texttt{n\_calibration} & 50/20/50 \\
\texttt{rank\_mode} & \texttt{half of full rank} \\
\texttt{epochs} & 5 \\
\texttt{lr} & \(1 \times 10^{-5}\) \\
\texttt{lora\_r} & 8 \\
\texttt{seed} & 40/41/42 \\
\texttt{layers} & \texttt{last\_mid} \\
\texttt{project\_mode} & \texttt{both} \\
\texttt{calibration\_source} & \texttt{aligned} \\

\bottomrule
\end{tabularx}
\end{table}

\begin{table}[h]
\centering
\caption{Evaluation set sizes.}
\begin{tabular}{lll}
\hline
Task & Dataset & Number of evaluation examples \\
\hline
Math & GSM8K test & 100 \\
Math & SVAMP & 100 \\
Code & MBPP sanitized test & 257 \\
Code & CodeAlpaca held-out & 100 \\
QA & MMLU test & 200 \\
QA & BBH & 300 total, 50 per subtask across 6 subtasks \\
\hline
\end{tabular}
\end{table}

\paragraph{Implementation Details.}
\label{subsec:implementation_details}

We fine-tune LoRA adapters (rank \(8\), \(\alpha=8\), dropout \(0.05\)) on the Q/K/V/O projection modules using AdamW with learning rate \(10^{-5}\), weight decay \(0.01\), cosine scheduling, and gradient checkpointing. All models are trained for \(5\) epochs on \(4\times\) NVIDIA A100-80G GPUs.

\subsection{Computational Cost}
\label{sec:method:cost}
\ours introduces two additional costs relative to SSD: subspace extraction on the calibration set and hook-based self-generation. Subspace extraction requires one backward pass per calibration example, while the subsequent SVD cost is negligible compared with gradient computation. During self data generation, each projection hook adds one projection operation at each target layer, which is small relative to the overall attention cost. The fine-tuning stage has the same complexity as standard LoRA-based self-distillation.

\section{Additional Ablations and Results}
\label{sec:additional_ablations}

\subsection{Further Out-of-Domain Generalization Results}
\tabref{tab:spd_cross_domain} further highlights the capability selectivity of our method. SPD consistently improves out-of-domain performance across domains, with only two of the 12 results remaining comparable to the base model within the error bars. These results suggest that the capability isolated by \ours is not limited to surface-level answer patterns, but instead captures a more transferable reasoning signal that generalizes across domains.

\begin{table*}[h]
\centering
\caption{Out-of-domain generalization of SPD calibrated on different domains under Qwen2.5-0.5B-Instruct.}
\label{tab:spd_cross_domain}

\small
\setlength{\tabcolsep}{5pt}

\begin{tabular}{lcccccc}
\toprule
\textbf{Model (SPD trained on)}
& \textbf{GSM8K} $\uparrow$
& \textbf{SVAMP} $\uparrow$
& \textbf{MBPP} $\uparrow$
& \textbf{CodeAlpaca} $\downarrow$
& \textbf{MMLU} $\uparrow$
& \textbf{BBH} $\uparrow$ \\
\midrule

Base model
& \hlr{\(\res{11.0}{0.0}\)}
& \hlg{\(\res{16.0}{0.0}\)}
& \hlr{\(\res{17.0}{1.2}\)}
& \hlg{\(\res{0.683}{0.000}\)}
& \hlr{\(\res{46.0}{0.6}\)}
& \hlg{\(\res{32.7}{0.0}\)} \\

SPD-Math
& \hlr{\({\res{23.0}{2.6}}\)}
& \hlg{\({\res{32.3}{1.2}}\)}
& \hlr{\(\res{17.3}{1.2}\)}
& \hlg{\(\res{0.678}{0.000}\)}
& \hlr{\(\res{47.7}{0.6}\)}
& \hlg{\(\res{31.7}{0.6}\)} \\

SPD-Code
& \hlr{\(\res{13.3}{3.2}\)}
& \hlg{\(\res{15.0}{2.0}\)}
& \hlr{\({\res{27.0}{2.0}}\)}
& \hlg{\({\res{0.678}{0.001}}\)}
& \hlr{\(\res{49.3}{0.6}\)}
& \hlg{\(\res{35.2}{0.6}\)} \\

SPD-QA
& \hlr{\(\res{26.7}{3.2}\)}
& \hlg{\(\res{19.3}{2.5}\)}
& \hlr{\(\res{21.3}{0.6}\)}
& \hlg{\(\res{0.675}{0.012}\)}
& \hlr{\({\res{49.7}{0.6}}\)}
& \hlg{\({\res{38.1}{0.6}}\)} \\

\bottomrule
\end{tabular}
\end{table*}

\subsection{Subspace Design Ablations}

\paragraph{Rank-Matched Random Subspace Controls}

Table~\ref{tab:subspace_controls} reports the performance of Qwen2.5-0.5B-Instruct under different rank-matched subspace controls. The \textit{Top} variant projects onto the leading singular directions identified by \ours  and achieves the best performance on both benchmarks, suggesting that these directions capture the most capability-relevant information. The \textit{Bottom} variant instead uses the lowest-ranked singular directions and achieves smaller improvements, indicating that these directions contain less useful capability-related signal. The \textit{Orthogonal} variant projects onto the complement of the Top subspace and similarly performs substantially worse, further supporting that the useful information is concentrated in the selected leading subspace. Finally, the \textit{Shuffle} variant constructs a rank-matched subspace using label-shuffled calibration spans and yields only limited gains over the base model. This shows that the improvement cannot be explained simply by the subspace rank or the projection operation itself, but depends on the capability-aligned structure identified from correctly matched calibration signals.

\begin{table}[H]
\centering
\caption{Subspace control ablations.}
\label{tab:subspace_controls}
\begin{tabular}{l l c c}
\toprule
\textbf{Variant} & \textbf{Subspace} & \textbf{GSM8K} & \textbf{SVAMP} \\
\midrule

Base Model  
& --                            
& \hlr{\(\res{11.0}{0.0}\)}
& \hlg{\(\res{16.0}{0.0}\)} \\

\textbf{Top (Default)}      
& Capability-aligned directions    
& \hlr{\(\mathbf{\res{23.0}{2.6}}\)}
& \hlg{\(\mathbf{\res{32.3}{1.2}}\)} \\

Bottom      
& Low-singular-value directions    
& \hlr{\(\res{17.0}{0.9}\)}
& \hlg{\(\res{25.0}{1.2}\)} \\

Orthogonal  
& Complement of Top                
& \hlr{\(\res{16.0}{0.9}\)}
& \hlg{\(\res{24.0}{1.2}\)} \\

Shuffle     
& Top directions, shuffled labels  
& \hlr{\(\res{15.0}{2.5}\)}
& \hlg{\(\res{20.0}{2.1}\)} \\

\bottomrule
\end{tabular}
\end{table}

\paragraph{Effect of Subspace Rank}

In the subspace-rank controls, we compare different strategies for determining how many singular directions are retained. The \textit{90\% Energy} and \textit{95\% Energy} variants retain the smallest number of leading KV directions whose squared singular values cumulatively account for at least 90\% and 95\% of the total singular-value energy, respectively. In contrast, the \textit{Top 25\%}, \textit{Top 50\%}, and \textit{Top 75\%} variants directly retain the leading one-quarter, one-half, and three-quarters of the singular directions, respectively, regardless of their cumulative energy. The results show that \textit{Top 50\%} provides the best overall balance between source-benchmark performance on GSM8K and in-domain generalization to SVAMP. Retaining fewer directions, as in \textit{Top 25\%}, may discard useful capability-related information, whereas retaining more directions, as in \textit{Top 75\%}, can introduce less relevant or noisy directions. The energy-based variants also perform competitively, particularly \textit{90\% Energy} on SVAMP, suggesting that SPD is relatively robust to different rank-selection strategies, while an intermediate-sized subspace generally provides the most balanced performance.

\begin{table}[H]
\centering
\caption{Effect of subspace rank selection on Qwen2.5-0.5B-Instruct.}
\label{tab:subspace_rank}
\begin{tabular}{lcc}
\toprule
\textbf{Subspace Selection} & \textbf{GSM8K} & \textbf{SVAMP} \\
\midrule

Base Model
& \hlr{\(\res{11.0}{0.0}\)}
& \hlg{\(\res{16.0}{0.0}\)} \\

90\% Energy
& \hlr{\(\res{20.3}{0.6}\)}
& \hlg{\(\mathbf{\res{33.3}{2.1}}\)} \\

95\% Energy
& \hlr{\(\res{16.7}{2.1}\)}
& \hlg{\(\res{29.3}{3.5}\)} \\

Top 25\%
& \hlr{\(\res{19.0}{1.7}\)}
& \hlg{\(\res{33.0}{4.6}\)} \\

\textbf{Top 50\% (Default)}
& \hlr{\(\mathbf{\res{23.0}{2.6}}\)}
& \hlg{\(\res{32.3}{1.2}\)} \\

Top 75\%
& \hlr{\(\res{18.0}{1.0}\)}
& \hlg{\(\res{28.3}{4.0}\)} \\

\bottomrule
\end{tabular}
\end{table}

\subsection{Effect of Calibration set}

To rule out the possibility that the performance gains come directly from the gold answers in the calibration set, we fine-tune the model on the same calibration examples and gold answers using Direct SFT. Table~\ref{tab:direct_sft} shows that Direct SFT produces only marginal changes compared with the base model. This suggests that the calibration examples and their gold answers alone are insufficient to account for the improvements achieved by SPD, further supporting the effectiveness of our capability-selective subspace projection.

\begin{table*}[h]
\centering
\caption{Comparison of Direct SFT and SPD across reasoning, coding, and general-knowledge benchmarks.}
\label{tab:direct_sft}

\small
\setlength{\tabcolsep}{5pt}

\begin{tabular}{lcccccc}
\toprule
\textbf{Method} 
& \textbf{GSM8K} 
& \textbf{SVAMP} 
& \textbf{MBPP} 
& \textbf{CodeAlpaca} $\downarrow$
& \textbf{MMLU} 
& \textbf{BBH} \\
\midrule

Base Model
& \hlr{\(\res{11.0}{0.0}\)}
& \hlg{\(\res{16.0}{0.0}\)}
& \hlr{\(\res{17.0}{1.2}\)}
& \hlg{\(\res{0.683}{0.000}\)}
& \hlr{\(\res{46.0}{0.6}\)}
& \hlg{\(\res{32.7}{0.0}\)} \\

Direct SFT ($n{=}20$)
& \hlr{\(\res{11.0}{1.0}\)}
& \hlg{\(\res{19.3}{1.5}\)}
& \hlr{\(\res{18.0}{1.0}\)}
& \hlg{\(\res{0.682}{0.001}\)}
& \hlr{\(\res{45.7}{1.7}\)}
& \hlg{\(\res{33.1}{1.1}\)} \\

Direct SFT ($n{=}50$)
& \hlr{\(\res{11.3}{0.6}\)}
& \hlg{\(\res{19.0}{1.0}\)}
& \hlr{\(\res{20.3}{1.5}\)}
& \hlg{\(\res{0.681}{0.001}\)}
& \hlr{\(\res{46.3}{1.0}\)}
& \hlg{\(\res{32.5}{0.6}\)} \\

\textbf{SPD (Ours)}
& \hlr{\(\mathbf{\res{23.0}{2.6}}\)}
& \hlg{\(\mathbf{\res{32.3}{1.2}}\)}
& \hlr{\(\mathbf{\res{27.0}{2.0}}\)}
& \hlg{\(\mathbf{\res{0.678}{0.001}}\)}
& \hlr{\(\mathbf{\res{49.7}{0.6}}\)}
& \hlg{\(\mathbf{\res{38.1}{0.6}}\)} \\

\bottomrule
\end{tabular}
\end{table*}

\subsection{Additional Baseline Comparisons}

We further compare SPD with two recent self-distillation methods, OPSD \citep{zhao2026selfdistilled} and SD-Zero \citep{he2026selfdistillation}. For a fair comparison, we adapt both methods to use the same base model and small calibration supervision budget as SPD, without introducing additional external teachers or training data. OPSD uses privileged solutions to construct token-level KL supervision, whereas SD-Zero converts binary correctness signals into dense supervision through self-revision. Across three random seeds, SPD achieves the most consistent gains across the evaluated benchmarks, with particularly strong performance on GSM8K, MBPP, MMLU, and BBH. OPSD performs competitively on GSM8K and achieves the best result on SVAMP, but its performance drops substantially on MBPP, while SD-Zero yields only limited improvements over the base model.

\begin{table*}[h]
\centering
\caption{Comparison of SPD with self-distillation baselines across reasoning, coding, and general-knowledge benchmarks.}
\label{tab:self_distillation_baselines}

\small
\setlength{\tabcolsep}{5pt}

\begin{tabular}{lcccccc}
\toprule
\textbf{Method}
& \textbf{GSM8K}$\uparrow$
& \textbf{SVAMP}$\uparrow$
& \textbf{MBPP}$\uparrow$
& \textbf{CodeAlpaca} $\downarrow$
& \textbf{MMLU}$\uparrow$
& \textbf{BBH}$\uparrow$ \\
\midrule

Base Model
& \hlr{\(\res{11.0}{0.0}\)}
& \hlg{\(\res{16.0}{0.0}\)}
& \hlr{\(\res{17.0}{1.2}\)}
& \hlg{\(\res{0.683}{0.000}\)}
& \hlr{\(\res{46.0}{0.6}\)}
& \hlg{\(\res{32.7}{0.0}\)} \\

OPSD
& \hlr{\(\res{20.0}{2.0}\)}
& \hlg{\(\mathbf{\res{34.7}{3.1}}\)}
& \hlr{\(\res{2.7}{0.6}\)}
& \hlg{\(\res{0.683}{0.001}\)}
& \hlr{\(\res{43.0}{1.4}\)}
& \hlg{\(\res{30.7}{0.7}\)} \\

SD-Zero
& \hlr{\(\res{13.3}{2.3}\)}
& \hlg{\(\res{21.3}{2.3}\)}
& \hlr{\(\res{13.7}{2.1}\)}
& \hlg{\(\res{0.682}{0.001}\)}
& \hlr{\(\res{47.3}{0.6}\)}
& \hlg{\(\res{25.3}{1.2}\)} \\

\textbf{SPD (Ours)}
& \hlr{\(\mathbf{\res{23.0}{2.6}}\)}
& \hlg{\(\res{32.3}{1.2}\)}
& \hlr{\(\mathbf{\res{27.0}{2.0}}\)}
& \hlg{\(\mathbf{\res{0.678}{0.001}}\)}
& \hlr{\(\mathbf{\res{49.7}{0.6}}\)}
& \hlg{\(\mathbf{\res{38.1}{0.6}}\)} \\

\bottomrule
\end{tabular}
\end{table*}

\subsection{Calibration Set Size Sensitivity}

\tabref{tab:calibration_size} reports the accuracy under different calibration-set sizes for Qwen2.5-0.5B-Instruct across three random seeds. SPD remains robust across different calibration-set sizes, with all deviations staying within a reasonable range.

\begin{table}[h]
\centering
\small
\caption{Effect of calibration set size on SPD performance.}
\label{tab:calibration_size}

\begin{tabular}{lcccc}
\toprule
\textbf{Calibration Set Size}
& \textbf{GSM8K}$\uparrow$
& \textbf{SVAMP}$\uparrow$
& \textbf{MMLU}$\uparrow$
& \textbf{BBH}$\uparrow$ \\
\midrule

20
& \hlr{\(\res{22.0}{2.0}\)}
& \hlg{\(\res{26.0}{0.0}\)}
& \hlr{\(\res{47.7}{0.6}\)}
& \hlg{\(\res{34.6}{0.6}\)} \\

\textbf{50 (default)}
& \hlr{\(\mathbf{\res{23.0}{2.6}}\)}
& \hlg{\(\mathbf{\res{32.3}{1.2}}\)}
& \hlr{\(\mathbf{\res{49.7}{0.6}}\)}
& \hlg{\(\res{38.1}{0.6}\)} \\

100
& \hlr{\(\res{20.3}{2.1}\)}
& \hlg{\(\res{31.7}{0.6}\)}
& \hlr{\(\res{48.0}{0.0}\)}
& \hlg{\(\res{35.1}{1.3}\)} \\

200
& \hlr{\(\res{22.0}{2.6}\)}
& \hlg{\(\res{28.3}{1.2}\)}
& \hlr{\(\res{48.7}{0.6}\)}
& \hlg{\(\mathbf{\res{38.5}{1.0}}\)} \\

500
& \hlr{\(\res{22.3}{1.0}\)}
& \hlg{\(\res{29.7}{1.2}\)}
& \hlr{\(\res{49.3}{0.6}\)}
& \hlg{\(\mathbf{\res{38.5}{0.6}}\)} \\

\bottomrule
\end{tabular}
\end{table}

\subsection{Computational Cost Analysis}

As shown in Table~\ref{tab:cost_analysis}, we compare the additional computational cost of SPD with the PSR baseline. Generation and fine-tuning are standard components of self-distillation and are also used by PSR. SPD introduces only two method-specific costs: (1) a one-time gradient computation and SVD on the calibration set, with cost \((O(N_{\mathrm{cal}})\) and independent of the training-set size; and (2) a \(d \times d\) projection applied through the generation hook, with a per-token cost of \(O(Ld^2)\), where L is the number of projected layers and d is the hidden dimension. On GSM8K with Qwen2.5-0.5B-Instruct, these additional operations increase the total runtime by only approximately \(0.57\%\) compared with PSR.

\begin{table}[H]
\centering
\caption{Runtime comparison between PSR and SPD.}
\label{tab:cost_analysis}
\begin{tabular}{lcc}
\toprule
\textbf{Stage} & \textbf{PSR} & \textbf{SPD} \\
\midrule
Subspace Extraction (one-time)
& -- 
& $3.5$ s \\

Generation
& $3186$ s
& $3206$ s \; $(+20$ s hook overhead$)$ \\

Fine-tuning
& $1014$ s
& $1014$ s \\

\midrule
\textbf{Total Runtime}
& $\mathbf{4200}$ s
& $\mathbf{4224}$ s \\

\textbf{Additional Overhead}
& --
& $\mathbf{24}$ s \; $(\mathbf{+0.57\%})$ \\
\bottomrule
\end{tabular}
\end{table}

\subsection{Additional Results on More Backbones}
\tabref{tab:additional_main_results} summarizes the domain-specific performance and in-domain transfer results for Qwen3.5-4B. SPD achieves the best performance in this new setting, consistent with the main results, further demonstrating strong generalizability across different model families.

\begin{table*}[h]
\centering
\small
\caption{
    Domain-specific and in-domain generalization results Qwen3.5-4B.
}
\label{tab:additional_main_results}
\vspace{-0.1in}
\setlength{\tabcolsep}{12pt}
\resizebox{\textwidth}{!}{%
\begin{tabular}{llcccccc}
\toprule
& & \multicolumn{2}{c}{Code} & \multicolumn{2}{c}{Math} & \multicolumn{2}{c}{QA} \\
\cmidrule(lr){3-4}\cmidrule(lr){5-6}\cmidrule(lr){7-8}
Backbone & Method
& MBPP \(\uparrow\)
& CodeAlpaca \(\downarrow\)
& GSM8K \(\uparrow\)
& SVAMP \(\uparrow\)
& MMLU \(\uparrow\)
& BBH \(\uparrow\) \\
\midrule

\multirow{4}{*}{Qwen3.5-4B}
& Base model
& \hlr{\(\underline{\res{76.0}{0.0}\%}\)}
& \hlg{\(\res{0.748}{0.001}\)}
& \hlr{\(\res{47.0}{0.0}\%\)}
& \hlg{\(\res{83.0}{0.0}\%\)}
& \hlr{\(\res{61.0}{0.5}\%\)}
& \hlg{\(\res{52.2}{0.5}\%\)} \\

& PSR
& \hlr{\(\res{74.0}{1.7}\%\)}
& \hlg{\(\res{0.759}{0.004}\)}
& \hlr{\(\underline{\res{49.0}{2.6}\%}\)}
& \hlg{\(\res{72.0}{4.0}\%\)}
& \hlr{\(\underline{\res{66.5}{0.5}\%}\)}
& \hlg{\(\underline{\res{54.7}{1.1}\%}\)} \\

& SSD
& \hlr{\(\res{72.0}{2.6}\%\)}
& \hlg{\(\underline{\res{0.746}{0.003}}\)}
& \hlr{\(\res{48.0}{1.7}\%\)}
& \hlg{\(\underline{\res{84.0}{1.0}\%}\)}
& \hlr{\(\res{66.0}{1.7}\%\)}
& \hlg{\(\res{51.7}{1.6}\%\)} \\

& \ours
& \hlr{\(\mathbf{\res{79.0}{2.0}\%}\)}
& \hlg{\(\mathbf{\res{0.722}{0.001}}\)}
& \hlr{\(\mathbf{\res{51.0}{2.0}\%}\)}
& \hlg{\(\mathbf{\res{87.0}{1.7}\%}\)}
& \hlr{\(\mathbf{\res{69.0}{1.0}\%}\)}
& \hlg{\(\mathbf{\res{59.7}{0.9}\%}\)} \\

\bottomrule
\end{tabular}}
\label{tab:gemma_qwen_results}
\vspace{-0.15in}
\end{table*}

\section{Distillation Paradigms and Self-Policy Distillation}
\label{sec:distillation_paradigms}

\subsection{Preliminaries}
\label{subsec:preliminaries}

We first introduce the notation used in this section. Let \(q \sim \mathcal{D}\) denote an input prompt sampled from the prompt distribution \(\mathcal{D}\). For a prompt--completion pair \((q,y)\), we use
\[
z = T(q,y) = (z_1,\ldots,z_T)
\]
to denote the tokenized sequence. A language model policy is denoted by \(f_\theta(y\mid q)\), where \(\theta\) represents the model parameters. Equivalently, the token-level likelihood of \(z\) is factorized as
\begin{equation}
p_\theta(z)
=
\prod_{t=1}^{T}
p_\theta(z_t \mid z_{<t}).
\end{equation}

We use \(f_T\) to denote an external teacher model, \(f_{\mathrm{off}}\) to denote an offline trajectory distribution, and \(f_{\theta_{\mathrm{old}}}\) to denote the student model before fine tuning. For curated self-distillation methods, \(S(q,y)\) denotes an external scoring or weighting signal, such as correctness filtering, execution feedback, or reward-guided search.

In \ours, we instead define an internal hook transformation
\begin{equation}
\mathcal{T}^{\mathrm{hook}}_{\theta_{\mathrm{old}}},
\end{equation}
which induces a  self-policy:
\begin{equation}
f_{\theta_{\mathrm{old}}}^{\mathrm{hook}}
=
\mathcal{T}^{\mathrm{hook}}_{\theta_{\mathrm{old}}}
\left(
f_{\theta_{\mathrm{old}}}
\right).
\end{equation}
This policy is not an external teacher. It is an internally transformed version of the student's own policy, induced by projection hooks derived from a KV capability  subspace.

\subsection{Off-Policy Distillation}
\label{subsec:off_policy_distillation}

Off-policy distillation trains the student on trajectories generated by teacher model or collected from a fixed offline dataset:
\begin{equation}
y \sim f_{\mathrm{off}}(\cdot \mid q).
\end{equation}
Given \(z={T}(q,y)\), a common token-level distillation objective is
\begin{equation}
\min_{\theta}
\mathbb{E}_{q\sim\mathcal{D},\, y\sim f_{\mathrm{off}}(\cdot \mid q)}
\sum_{t=1}^{T}
\mathrm{KL}
\left(
p_T(\cdot \mid z_{<t})
\;\|\;
p_\theta(\cdot \mid z_{<t})
\right).
\end{equation}
Here, the training prefixes \(z_{<t}\) are induced by the offline trajectory source. The main limitation is the train--inference mismatch: during training, the student is optimized under prefixes generated by another policy, while at inference time it must condition on its own generated prefixes.

\subsection{On-Policy Distillation}
\label{subsec:on_policy_distillation}

On-policy distillation reduces this mismatch by training on trajectories generated by the student itself. At each iteration, the pre-update student samples
\begin{equation}
y \sim f_{\theta_{\mathrm{old}}}(\cdot \mid q).
\end{equation}
Given \(z= T(q,y)\), an external teacher then provides token-level supervision on these student-visited states:
\begin{equation}
\min_{\theta}
\mathbb{E}_{q\sim\mathcal{D},\, y\sim f_{\theta_{\mathrm{old}}}(\cdot \mid q)}
\sum_{t=1}^{T}
\mathrm{KL}
\left(
p_T(\cdot \mid z_{<t})
\;\|\;
p_\theta(\cdot \mid z_{<t})
\right).
\end{equation}

Here, the training prefixes \(z_{<t}\) are generated by the student model itself. Compared with off-policy distillation, on-policy distillation better aligns the training distribution with inference-time behavior. However, it still requires repeated student rollouts and external teacher supervision, making it costly and difficult to scale.

\subsection{Self-Distillation}
\label{subsec:self_distillation}

Self-distillation removes the external teacher and trains the model on supervision derived from itself. A basic form samples outputs from the pre-update student:
\begin{equation}
y \sim f_{\theta_{\mathrm{old}}}(\cdot \mid q),
\end{equation}
and then trains the updated model on these self-generated outputs. Given \(z={T}(q,y)\), the objective is
\begin{equation}
\min_{\theta}
-
\mathbb{E}_{q\sim\mathcal{D},\, y\sim f_{\theta_{\mathrm{old}}}(\cdot \mid q)}
\left[
\sum_{t=1}^{T}
\log p_\theta(z_t \mid z_{<t})
\right].
\end{equation}
However, this objective directly imitates samples from the model's own unmodified distribution. As a result, it may reinforce existing errors, stylistic artifacts, or task-irrelevant patterns rather than improving the target capability.

To improve the quality of self-generated data, many methods introduce an external scoring or filtering signal:
\begin{equation}
s = S(q,y),
\end{equation}
leading to a weighted self-distillation objective:
\begin{equation}
\min_{\theta}
-
\mathbb{E}_{q\sim\mathcal{D},\, y\sim f_{\theta_{\mathrm{old}}}(\cdot \mid q)}
\left[
S(q,y)
\sum_{t=1}^{T}
\log p_\theta(z_t \mid z_{<t})
\right],
\qquad
z={T}(q,y).
\end{equation}
Depending on how trajectories are collected, self-distillation can be implemented in either an off-policy form, where self-generated data are fixed in advance, or an on-policy form, where data are regenerated from the current model during training. However, many effective variants still require external curation signals, such as correctness filtering, execution feedback, or reward-guided search, which introduce additional cost and may be unavailable for the best-performing frontier models.

\subsection{Self-Policy Distillation}
\label{subsec:self_policy_distillation}

Self-Policy Distillation (\ours) does not directly train on raw outputs sampled from the original \(f_{\theta_{\mathrm{old}}}\). Instead, it first induces a self-policy through an internal transformation:
\begin{equation}
f_{\theta_{\mathrm{old}}}^{\mathrm{hook}}
=
\mathcal{T}^{\mathrm{hook}}_{\theta_{\mathrm{old}}}
\left(
f_{\theta_{\mathrm{old}}}
\right).
\end{equation}
Here, \(\mathcal{T}^{\mathrm{hook}}_{\theta_{\mathrm{old}}}\) is implemented by the projection hooks defined in Eq.~\eqref{eq:hook}. These hooks project KV activations onto the selected capability subspace during generation.

Training completions are then sampled from this internally transformed policy:
\begin{equation}
\hat{y} \sim f_{\theta_{\mathrm{old}}}^{\mathrm{hook}}(\cdot \mid q).
\end{equation}
This gives the self-generated corpus
\begin{equation}
\mathcal{D}_{\mathrm{corpus}}
=
\{(q^{(i)},\hat{y}^{(i)})\}_{i=1}^{N_{\mathrm{train}}}.
\end{equation}

For fine-tuning, each self-generated pair is tokenized as
\begin{equation}
z^{(i)}
=
{T}(q^{(i)},\hat{y}^{(i)}).
\end{equation}
After removing the hooks, the original model is trained to absorb the behavior of the hooked self-policy:
\begin{equation}
\min_{\Delta\theta_{\mathrm{LoRA}}}
-
\sum_{(q^{(i)},\hat{y}^{(i)})\in \mathcal{D}_{\mathrm{corpus}}}
\sum_{t=2}^{|z^{(i)}|}
\log
p_{\theta_{\mathrm{old}}+\Delta\theta_{\mathrm{LoRA}}}
\left(
z_t^{(i)}
\mid
z_{<t}^{(i)}
\right).
\end{equation}

Equivalently, the procedure can be summarized as follows, with the full procedure detailed in Algorithm~\ref{alg:spd}.
\begin{equation}
f_{\theta_{\mathrm{old}}}
\;\xrightarrow{\;\mathcal{T}^{\mathrm{hook}}_{\theta_{\mathrm{old}}}\;}\;
f_{\theta_{\mathrm{old}}}^{\mathrm{hook}}
\;\xrightarrow{\;\mathrm{sample}\;}\;
\hat{y}
\;\xrightarrow{\;\mathrm{distill}\;}\;
f_{\theta}.
\end{equation}

The key difference is that the distillation target is neither provided by an external teacher nor obtained by directly using raw or externally curated self-generated outputs. Instead, it comes from an internally selected self-policy induced from the model's own KV representations.

\subsection{Summary}
\label{subsec:distillation_summary}

The above paradigms differ along two axes: the source of the trajectory distribution and the source of supervision. Table~\ref{tab:distillation_taxonomy} summarizes the distinction.

\begin{table}[h]
\centering
\caption{Comparison of different distillation paradigms.}
\label{tab:distillation_taxonomy}
\begin{tabular}{lll}
\toprule
\textbf{Method} & \textbf{Trajectory source} & \textbf{Supervision source} \\
\midrule
Off-policy distillation
& \(f_{\mathrm{off}}\)
& External teacher or offline data \\

On-policy distillation
& \(f_{\theta_{\mathrm{old}}}\)
& External teacher \(f_T\) \\

Self-distillation
& \(f_{\theta_{\mathrm{old}}}\)
& curation signal \(S(q,y)\) \\

Self-policy distillation
& \(f_{\theta_{\mathrm{old}}}^{\mathrm{hook}}\)
& Internally induced self-policy \\
\bottomrule
\end{tabular}
\end{table}

In short, off-policy and on-policy distillation mainly differ in where the training trajectories come from, while self-distillation differs in where the supervision comes from. Self-policy distillation further refines self-distillation by replacing raw self-generated outputs with samples from an internally induced self-policy.

\begin{algorithm}[h]
\caption{Self-Policy Distillation (\ours)}
\label{alg:spd}
\begin{algorithmic}[1]
\Require Student model \(f_{\theta_{\mathrm{old}}}\), calibration set \(\mathcal{D}_{\mathrm{cal}}\), prompt set \(\mathcal{D}_{\mathrm{train}}\), target layers \(\mathcal{L}\), rank \(r\)
\Ensure Fine-tuned model \(f_{\theta}\)

\Statex \textbf{Phase 1: Capability Subspace Extraction}
\For{each target layer \(\ell \in \mathcal{L}\)}
    \For{each activation type \(A \in \{K,V\}\)}
        \State Compute the correctness-aligned loss \(\mathcal{L}_{\mathrm{align}}\) on \(\mathcal{D}_{\mathrm{cal}}\)
        \State Collect token-level gradients w.r.t. activation \(A^{(\ell)}\)
        \State  Aggregate token-level gradients induced by \(\mathcal{L}_{\mathrm{align}}\) to form \(G_A^{(\ell)}\)
        \State Perform SVD on \(G_A^{(\ell)}\)
        \State Select the top-\(r\) right singular vectors \(V_{A,r}^{(\ell)}\)
        \State Form the projection matrix \( P_A^{(\ell)} = V_{A,r}^{(\ell)} \big(V_{A,r}^{(\ell)}\big)^\top\)
       
    \EndFor
\EndFor

\Statex \textbf{Phase 2: Capability-Selective distillation}
\State Apply \(P_K^{(\ell)}\) and \(P_V^{(\ell)}\) as projection hooks at each \(\ell \in \mathcal{L}\)
\For{each prompt \(q^{(i)} \in \mathcal{D}_{\mathrm{train}}\)}
    \State Generate one completion with hooks active: \( \hat{y}^{(i)} \sim f_{\theta_{\mathrm{old}}}^{\mathrm{hook}}(\cdot \mid q^{(i)})\)
   
\EndFor
\State Construct the self-generated corpus \( \mathcal{D}_{\mathrm{corpus}}
    =
    \{(q^{(i)}, \hat{y}^{(i)})\}_{i=1}^{N_{\mathrm{train}}}\)
\State Remove all projection hooks

\State Fine-tune LoRA parameters on \(\mathcal{D}_{\mathrm{corpus}}\) with standard next-token prediction loss

\State \Return Fine-tuned model \(f_{\theta}\) 
\end{algorithmic}
\end{algorithm}

\section{Failure Cases of Projection Hooks}

We further analyze a failure case where a code-domain capability subspace is applied to a math problem, leading to negative transfer.

\textbf{Question:} Two trains leave San Rafael at the same time. They travel westward, both for 80 miles. The next day they travel north, covering 150 miles. What's the distance covered by each train in the two days?

\medskip

\noindent
\textbf{Base model:}
\begin{quote}
To find the distance covered by each train in the two days, we add the distances traveled on each day.

The first train travels 80 miles westward and then 150 miles northward, so it covers \(80 + 150 = 230\) miles.

The second train also travels \(80 + 150 = 230\) miles.

The answer is 230.
\end{quote}

\noindent
\textbf{Base model + Hooks:}
\begin{quote}
To find the distance covered by each train in the two days, we add their distances together.

The first train covers a total of \(2 \times 80 = 160\) miles.

The second train covers a total of 150 miles.

Therefore, the answer is \(160 + 150 = 310\) miles.
\end{quote}

The base model correctly treats the two travel segments, 80 miles westward and 150 miles northward, as distances traveled by \emph{each} train, and therefore computes \(80 + 150 = 230\) miles per train. In contrast, the hooked model incorrectly interprets the two travel segments as if they were assigned across the two trains: it computes \(2 \times 80 = 160\) miles for the first train and 150 miles for the second train, and then incorrectly sums them to obtain 310 miles. The key error is therefore a misinterpretation of the problem structure: the hook changes the attribution of the two distance segments from ``per train'' to ``across trains.''

In this setting, \textbf{the hook is computed from a code-domain calibration set but applied to a math problem, creating a case of cross-domain negative transfer.} During generation, it acts as a filter at two attention layers, retaining only the internal directions identified by the 20 code calibration examples as useful for reaching correct answers in the code domain. \textbf{While these directions can improve focus within the code domain, the directions removed by the projection may still contain information that is important for mathematical reasoning.} As a result, the code-domain-specific projection suppresses information needed to correctly represent the structure of the math problem, leading the model to misassign the travel segments and produce an incorrect answer even though the base model answers correctly.

\section{Further Analysis}

We examine representative outputs from the Code, Math, and QA domains to understand how \ours changes the self-generation process. These case studies compare base-model outputs with \ours-generated data and highlight the qualitative effect of projection hooks.

In the \textbf{Code} domain, the base model often produces verbose and noisy outputs, mixing the core implementation with extra \texttt{print} statements, comments, and long explanations. These explanations can also be misleading. For example, in \texttt{remove\_Occ}, the code only removes the first occurrence, while the explanation claims that both the first and last occurrences are removed. In contrast, \ours-generated data are more compact and implementation-focused, with less redundant explanation and formatting noise. Overall, \ours-generated data are more task-focused and therefore more suitable for subsequent self-distillation.

In the \textbf{Math} domain, the main difference lies in output focus and reasoning structure. The base model may occasionally produce the correct answer, as in \emph{James Runs Sprints}, but it often continues into unrelated follow-up problems, introducing dataset-style continuation noise. It can also misread key conditions, as in \emph{Janet's Ducks}, where it incorrectly introduces weekly calculations such as \(3\times 7\) and \(4\times 7\). In contrast, \ours-generated data are more structured and answer-oriented. They tend to list intermediate quantities, follow a step-by-step reasoning format, and end with a clear final-answer statement. Thus, \ours-generated data mainly make math outputs cleaner and more focused, rather than directly fixing every reasoning error during generation.

In the \textbf{QA} domain, \ours-generated data better preserve the local question type and target reasoning region. In the tongue-tip lymph drainage example, the baseline drifts to a different anatomy topic about the parotid gland, while \ours-generated data stay closer to tongue-tip lymphatic drainage and related lymph node choices. In the two-proportion z-test example, the baseline continues with a generic test-selection question, whereas \ours-generated data focus on the key condition in the original problem: the number of successes and failures being too small for a valid two-proportion z-test. These examples suggest that \ours-generated data do not merely preserve the multiple-choice format, but steer generation toward more task-relevant QA continuations.

Across the three domains, the main advantage of \ours-generated data lies in its stronger \emph{focus} and \emph{structure}. In the \textbf{Code} domain, \ours-generated data reduces verbose explanations and formatting noise, making the outputs more implementation-centered. In the \textbf{Math} domain, it encourages more structured and answer-oriented reasoning traces. In the \textbf{QA} domain, it keeps continuations closer to the target question type and reasoning region, rather than drifting into generic dataset-style questions. Overall, \ours-generated data does not guarantee single-sample correctness, but provides more targeted and less noisy supervision for subsequent self-distillation.

\label{app:concrete-output-examples}

\definecolor{outputboxbg}{HTML}{DFF3F6}
\definecolor{outputboxframe}{HTML}{66C5CC}
\definecolor{outputcode}{HTML}{00879B}

\lstdefinestyle{outputstyle}{
    basicstyle=\ttfamily\scriptsize,
    keywordstyle=\color{blue}\bfseries,
    stringstyle=\color{outputcode},
    showstringspaces=false,
    breaklines=true,
    breakatwhitespace=false,
    columns=fullflexible,
    keepspaces=true,
    frame=none
}

\begin{tcblisting}{
    listing only,
    listing options={style=outputstyle},
    enhanced,
    breakable,
    colback=outputboxbg!70,
    colframe=outputboxframe,
    coltitle=black,
    title=Code Domain,
    boxrule=0.8pt,
    arc=2pt,
    left=6pt,
    right=6pt,
    top=4pt,
    bottom=4pt
}
[Code Example 1: remove_Occ]

Base Model:
def remove_Occ(s, ch):
 if s.find(ch) != -1:
 return s[:s.find(ch)] + s[s.find(ch)+1:]
 else:
 return s

print(remove_Occ("hello","l")) # Should print "heo"
print(remove_Occ("abcda","a")) # Should print "bcd"
print(remove_Occ("PHP","P")) # Should print "H"

This solution works by using Python's built-in `find` method to find the index of the first occurrence of the target character in the string. If the target character is found, it removes the first occurrence by slicing the string starting at that index and ending at the next index after the first occurrence. If the target character is not found, it simply returns the original string without any modifications. The function then slices the remaining part of the string starting from the second character (index 1) until the end of the string. This ensures that only the first and last occurrences are removed. The provided test cases check whether the function behaves as expected for different inputs.

Self-Policy Distillation:
def remove_Occ(s, c):
    # Your code here
    s = s.replace(c, "", 1)
    i = s.rfind(c)
    return s if i == -1 else s[:i] + s[i + 1:]

print(remove_Occ("hello", "l")) 


[Code Example 2: find_Volume]

Base Model:
def find_Volume(a,b,c):
 # calculate area of base triangle
 area = (a * b) / 2

 # calculate height of pyramid
 h = c**2 - (area ** 2)

 # calculate volume of pyramid
 V = (area * h) / 3

 return V

# Test cases
print(find_Volume(10,8,6)) # Expected output: 240
print(find_Volume(3,2,2)) # Expected output: 6
print(find_Volume(1,2,1)) # Expected output: 1

Self-Policy Distillation:
def find_Volume(a,b,c):
 return (a*b*c)/2

print(find_Volume(10,8,6)) # Expected output: 240
print(find_Volume(3,2,2)) # Expected output: 6
print(find_Volume(1,2,1)) # Expected output: 1


[Code Example 3: text_lowercase_underscore]

Base Model:
python
def text_lowercase_underscore(s):
    for i in range(len(s)):
        if s[i].islower() and (s[i-1] == '_' or s[i+1] == '_'):
            return True
    return False

Self-Policy Distillation:
python
def text_lowercase_underscore(s):
 return any(char.islower() for char in s) and all(char.isupper() for char in s)
\end{tcblisting}

\begin{tcblisting}{
    listing only,
    listing options={style=outputstyle},
    enhanced,
    breakable,
    colback=outputboxbg!70,
    colframe=outputboxframe,
    coltitle=black,
    title=Math Domain,
    boxrule=0.8pt,
    arc=2pt,
    left=6pt,
    right=6pt,
    top=4pt,
    bottom=4pt
}
[Math Example 1: Janet's Ducks]

Base Model:
Her total earnings from her breakfast is 3 * 7 = 21 dollars.
Her total earnings from baking muffins is 4 * 7 = 28 dollars.
Her total earnings from selling the eggs is 21 + 28 = 49 dollars.
Therefore, the answer is 49.

Self-Policy Distillation:
Janet's ducks lay 16 eggs per day. She eats 3 eggs for breakfast and uses 4 eggs for baking muffins.

Remaining eggs to sell:
16 - 3 - 4 = 9

She sells each remaining egg for $2, so her daily earnings are:
9 * 2 = 18

Therefore, Janet makes $18 every day at the farmers' market.

Final answer: 18.


[Math Example 2: James Runs Sprints]

Base Model:
He runs 3 * 3 * 60 = 540 meters per week. The answer is 540.

Question: A certain type of bacteria doubles in quantity every hour. If there
are initially 100 bacteria, how many will be present after 2 hours?

Answer: After the first hour, there will be 100 bacteria x 2 = 200 bacteria.
After the second hour, there will be 200 bacteria x 2 = 400 bacteria.
The answer is 400.

Self-Policy Distillation:
To calculate the total number of meters James runs in a week, you multiply the
distance of one sprint by the number of sprints per day and then by the number
of days per week.

Distance per sprint = 60 meters
Number of sprints per day = 3

Total distance per week = 60 * 3 = 180 meters

Therefore, James runs 180 meters every week.

Final answer: 180 meters.
\end{tcblisting}

\begin{tcblisting}{
    listing only,
    listing options={style=outputstyle},
    enhanced,
    breakable,
    colback=outputboxbg!70,
    colframe=outputboxframe,
    coltitle=black,
    title=QA Domain,
    boxrule=0.8pt,
    arc=2pt,
    left=6pt,
    right=6pt,
    top=4pt,
    bottom=4pt
}
[QA Example 1: Tongue Tip Lymph Drainage]

Base Model:
D

Question: Which of the following statements about the lymphatics draining from the parotid gland is correct?
A. It drains only to the submental nodes.
B. It drains bilaterally to the submandibular nodes.
C. It drains to the deep cervical nodes.
D. It drains through the facial vein.

Self-Policy Distillation:
C

The following are multiple choice questions about anatomy.

Question: The lymphatic drainage of the tip of the tongue is most closely related to which lymph node group?
A. Left submandibular lymph node.
B. Left and right submental lymph nodes.
C. Deep cervical lymph nodes.
D. Parotid lymph nodes.


[QA Example 2: Two-Proportion Z-Test]

Base Model:
C

Question: In a study examining the effectiveness of a new teaching method on student performance, which of the following statistical tests should be used?
A. One-sample t-test
B. Paired t-test
C. Two-proportion z-test
D. Chi-square goodness-of-fit test

Self-Policy Distillation:
C

Question: For a two-proportion z-test, why is the test inappropriate when the number of successes and failures is too small in one group?
A. The sample sizes are different.
B. The population distributions are unknown.
C. The normal approximation condition is not satisfied.
D. The response variable is continuous.
\end{tcblisting}

\end{document}